\documentclass[sigconf]{acmart}
\usepackage{amsthm}
\usepackage{amsmath}
\usepackage{algorithm}
\usepackage{enumitem}
\usepackage{picinpar}
\usepackage{lineno}
\usepackage{graphicx}
\usepackage{placeins}
\usepackage{algorithmicx}
\usepackage[noend]{algpseudocode}
\usepackage{subfigure}
\usepackage{multirow}
\usepackage{color}
\usepackage{balance}
\usepackage{enumitem}
\usepackage{hhline}
\usepackage[normalem]{ulem}
\usepackage{booktabs}
\usepackage{wrapfig}
\usepackage{cancel}
\usepackage{placeins}
\usepackage{hyperref}
\usepackage{makecell}
\usepackage{booktabs}
\usepackage{threeparttable}
\usepackage{graphicx}
\usepackage[table]{xcolor}
\definecolor{oursgray}{gray}{0.92}
\usepackage{xcolor}
\usepackage{cleveref}

\usepackage[table]{xcolor}
\definecolor{oursgray}{RGB}{240,240,240}

\newcommand{\stitle}[1]{\vspace{1mm} \noindent {\bf #1}}
\usepackage{listings}

\lstdefinestyle{jsonsample}{
  basicstyle=\ttfamily\small,
  columns=fullflexible,
  breaklines=true,
  breakatwhitespace=true,
  showstringspaces=false,
  keepspaces=true,
  upquote=true,
  literate={°}{{\textdegree}}1
}

\newcommand{\method}[1]{\textsc{#1}}
\newcommand{\model}{\method{UrbanGraphEmbedding}{}}
\newcommand{\modelS}{\method{UGE}{}}
\newcommand{\benchmark}{\method{UGBench}{}}
\newcommand{\dataset}{\method{UGData}{}}
\newcommand{\NoEdgeEncoder}{\method{NoEdgeEncoder}{}}
\newcommand{\NoPosEncoder}{\method{NoPosEncoder}{}}
\newcommand{\NodeAttrEncoder}{\method{NodeAttrEncoder}{}}
\newcommand{\Stage}{\method{Stage1-only}{}}
\newcommand{\stage}{\method{Stage2-only}{}}
\usepackage{xcolor}
\newcommand{\best}[1]{\textbf{\textcolor{black}{#1}}}

\usepackage{listings}
\usepackage{xcolor}

\AtBeginDocument{%
  }

\setcopyright{acmlicensed}
\copyrightyear{2018}
\acmYear{2018}
\acmDOI{XXXXXXX.XXXXXXX}
\acmConference[Conference acronym 'XX]{Make sure to enter the correct
  conference title from your rights confirmation email}{June 03--05,
  2018}{Woodstock, NY}
\acmISBN{978-1-4503-XXXX-X/2018/06}

\begin{document}

\title[UrbanGraphEmbeddings]{UrbanGraphEmbeddings: Learning and Evaluating Spatially Grounded Vision–Language Embeddings for Urban Environments}

\author{Jie Zhang$^*$}
\affiliation{%
 \institution{National University of Singapore}
 \institution{Department of Architecture}
  \country{Singapore}}
\email{jiezhang_jz@u.nus.edu}

\author{Xingtong Yu$^*$}
\affiliation{%
  \institution{The Chinese University of Hong Kong}
  \department{Dept of Systems Eng. \& Eng. Mgmt.}
  \country{China}}
\email{xtyu@se.cuhk.edu.hk}

\author{Yuan Fang$^{\dagger}$}
\affiliation{%
  \institution{Singapore Management University}
  \department{School of Computing \& Info. Systems}
  \country{Singapore}}
\email{yfang@smu.edu.sg}

\author{Rudi Stouffs$^{\dagger}$}
\affiliation{%
  \institution{National University of Singapore}
  \department{Department of Architecture}
  \country{Singapore}}
\email{stouffs@nus.edu.sg}

\author{Zdravko Trivic$^{\dagger}$}
\affiliation{%
  \institution{National University of Singapore}
  \department{Department of Architecture}
  \country{Singapore}}
\email{akizt@nus.edu.sg}
\thanks{
    $^*$Co-first authors.\\
    $^{\dagger}$Corresponding authors.}

\renewcommand{\shortauthors}{Zhang et al.}

\begin{abstract}
Learning transferable multimodal embeddings for urban environments is challenging because urban understanding is inherently spatial, yet existing datasets and benchmarks lack explicit alignment between street-view images and urban structure. We introduce \dataset, a spatially grounded dataset that anchors street-view images to structured spatial graphs and provides graph-aligned supervision via spatial reasoning paths and spatial context captions, exposing distance, directionality, connectivity, and neighborhood context beyond image content. Building on \dataset, we propose \modelS, a two-stage training framework that progressively and stably aligns images, text, and spatial structures by combining instruction-guided contrastive learning with graph-based spatial encoding. We finally introduce \benchmark, a comprehensive benchmark to evaluate how spatially grounded embeddings support diverse urban understanding tasks---including geolocation ranking, image retrieval, urban perception, and spatial grounding. We develop \modelS\ on multiple state-of-the-art VLM backbones, including Qwen2-VL, Qwen2.5-VL, Phi-3-Vision, and LLaVA1.6-Mistral, and train fixed-dimensional spatial embeddings with LoRA tuning. \modelS\ built upon Qwen2.5-VL-7B backbone achieves up to 44\% improvement in image retrieval and 30\% in geolocation ranking on training cities, and over 30\% and 22\% gains respectively on held-out cities, demonstrating the effectiveness of explicit spatial grounding for spatially intensive urban tasks. The code and datasets are available at \textcolor{blue}{\url{https://github.com/Jaygagaga/UGE/tree/main}}. 

\end{abstract}

\begin{teaserfigure}
  \centering
  \includegraphics[width=\textwidth]{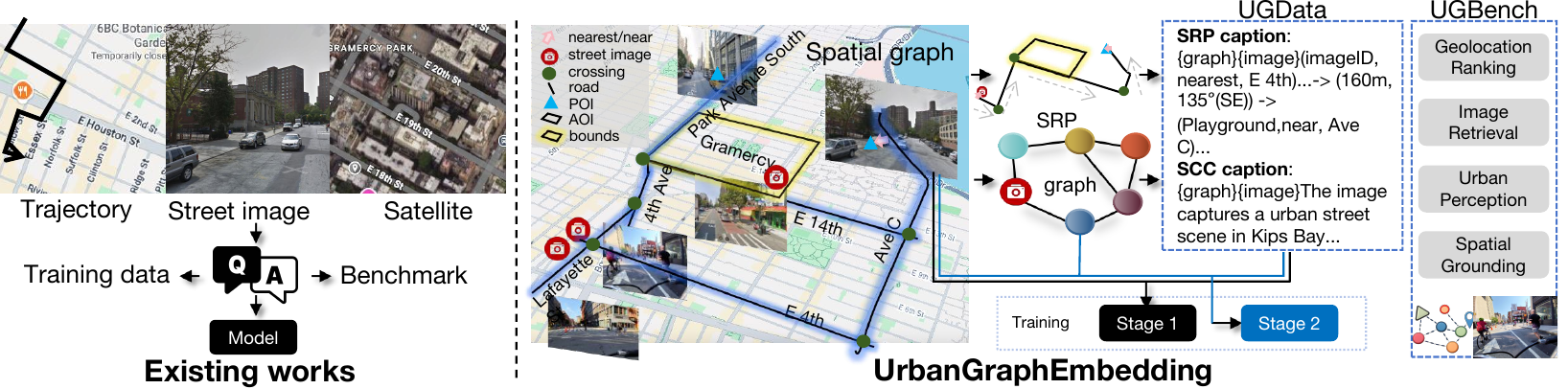}
  \vspace{-5mm}
  \caption{Comparison between existing urban VLM approaches and \model. Existing methods mainly rely on QA-style supervision and image--text alignment, whereas \model\  aligns images, text, and spatial graphs to learn and evaluate spatially grounded urban representations using UGData and UGBench.}
  \label{fig:motivation}
\end{teaserfigure}




\maketitle

\section{Introduction}


Urban science is increasingly powered by learning representations from multi-source urban observations, ranging from street-level imagery, remote sensing data, geo-text, to 3D point clouds and spatiotemporal trajectories. This has led to a growing reliance on vision–language models (VLMs) \cite{feng2025urbanllava,hao2025urbanvlp} as a general-purpose backbone for urban representation learning. By pre-training on large-scale image–text corpora to acquire transferable multimodal alignments, these models can be directly reused or lightly adapted to support a wide range of urban tasks, such as geolocation prediction \cite{haas2024pigeon, wu2022im2city, daruna2025geosurge, clark2023we}, socieconomic indicator prediction \cite{liu2025cityrise}, navigation prediction \cite{schumann2024velma,zhang2025citynavagent}, spatial reasoning \cite{feng2025urbanllava,xu2025geonav} and urban scene understanding \cite{randhawa2025paved,cordts2016cityscapes}.

However, urban understanding is not simply visual and textual, but spatial and relational. The semantics of a location is not solely determined by what is visible in a single street-view image or expressible in a local caption, but also by how the location is positioned within the city’s topology, such as its connectivity, proximity, and broader neighborhood structure \cite{ewing2009measuring,lynch1960image}. As shown in Fig.~\ref{fig:motivation}, current urban VLMs rely on instruction tuning and QA-style supervision, encoding spatial information implicitly through language and confining it to isolated question contexts, resulting in fragmented, task-specific representations that lack explicit urban grounding.
This limitation stems from the lack of datasets that explicitly align spatial knowledge with images and texts, leaving urban VLMs to optimize primarily for visual--textual alignment without an explicit pathway to encode geographic structure beyond the camera’s field of view. This gap motivates the creation of a benchmark that enhances urban VLM fine-tuning and evaluation by integrating urban spatial data to better capture spatial relationships. However, this is non-trival due to two key challenges.

First, \textit{how to extract spatial knowledge and align with street-level images, capturing urban relationships beyond what is directly visible?} Urban spatial knowledge is inherently structured, heterogeneous, relational, and multi-scale, and cannot be adequately captured by isolated image--text pairs. Moreover, effective supervision for spatial representation learning must be compatible with existing VLM training paradigms. To address this challenge at the data level, we leverage urban spatial graphs \cite{iddianozie2020improved,iwaniak2023spatial,yeghikyan2020learning}, which use low-cost, large-scale open geographic data to model locations as nodes connected by roads and functional relations. From such spatial graphs, we construct \textbf{\dataset}, a spatially grounded dataset that explicitly anchors each street-view image to its surrounding urban context. Specifically, for each image, we assign a localized spatial subgraph and derive structured, language-aligned supervision signals, including \emph{spatial reasoning paths} (SRPs) and \emph{spatial context captions} (SCCs),  describing how the image location is situated within the broader urban environment. \dataset~ pipeline is low-cost and highly scalable, with supervision grounded in reliable geographic data rather than manual annotation. Building on this pipeline, we treat spatial graph as a standalone input modality aligned with street-level images, enabling multimodal embedding learning that captures structured urban context beyond visual appearance alone.

Second, \emph{how can structured spatial knowledge be progressively injected into VLM embeddings in a stable and transferable manner, enabling general-purpose spatial representations?} Na\"ively combining images, text, and spatial graphs during training can lead to unstable optimization, as pretrained visual--language representations can be sensitive to distribution shifts introduced by an additional modality. Hence, we propose a two-stage training framework, \textbf{\model}, that progressively injects spatial knowledge into VLM embeddings while preserving their visual--language alignment. Specifically, spatial awareness is first induced via textual spatial reasoning cues in Stage 1, followed by explicit urban spatial structure encoding in Stage 2 through a graph encoder that propagates topological and relational information from spatial subgraphs.

To evaluate the learned representations, we introduce \textbf{\benchmark}, a benchmark that assesses performance across various urban science tasks: geolocation ranking, image retrieval, urban perception, and spatial grounding. It adopts a unified embedding-based ranking framework, supporting mono-modal or graph-augmented inputs under zero-shot evaluation.

In summary, our contributions are threefold:
(1) We introduce a spatially grounded urban dataset with spatial reasoning paths and spatial context captions that explicitly align street-view images with graph-structured urban context.
(2) We propose a two-stage training strategy that progressively injects structured spatial knowledge into VLM embeddings via instruction-guided contrastive learning and graph-conditioned encoding.
(3) We present a systematic benchmark that evaluates spatial representations in a zero-shot setting, assessing whether spatial context improves retrieval and ranking across diverse urban science tasks.

\begin{figure*}[!t]
    \centering
    \includegraphics[width=1\linewidth]{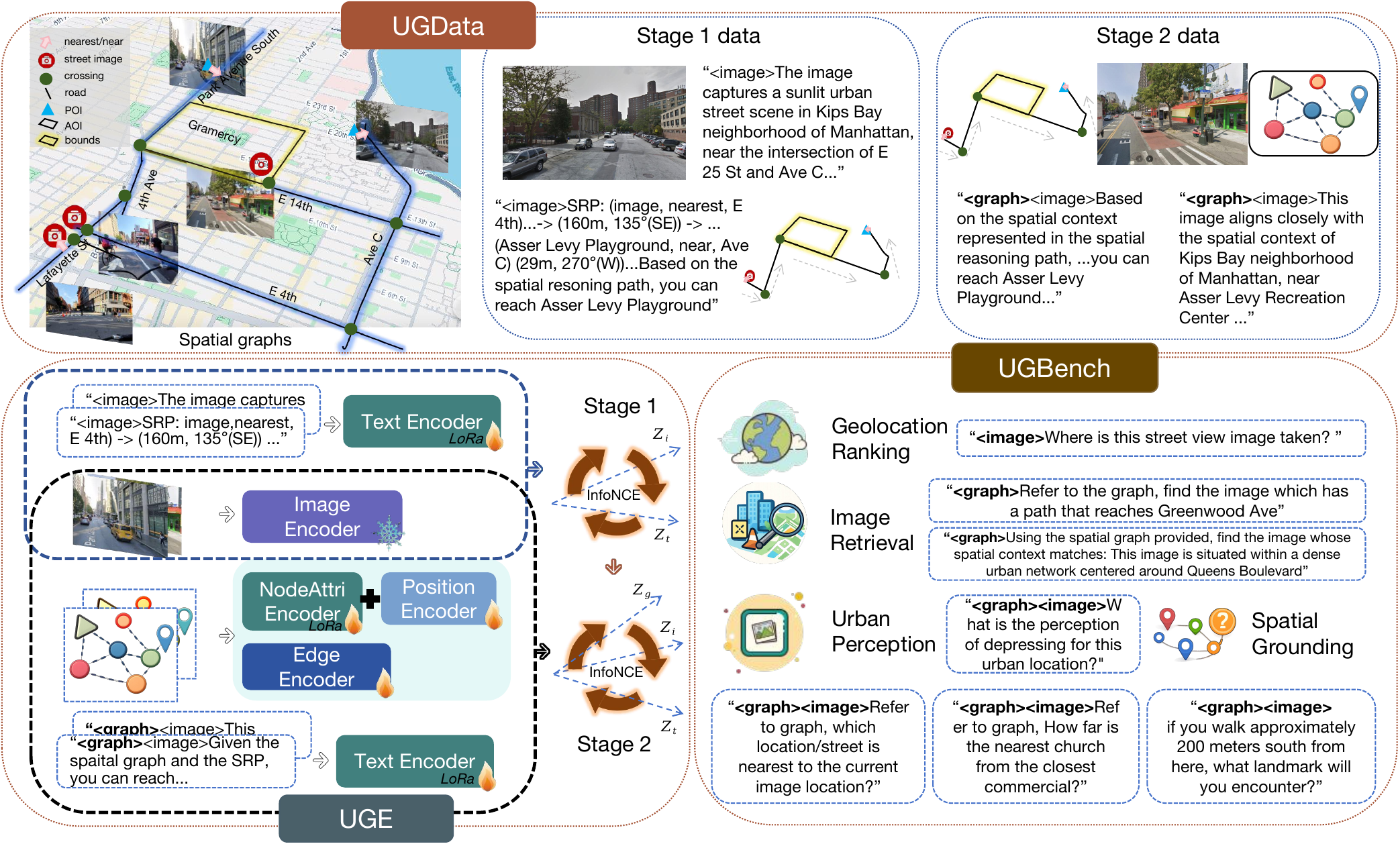}
    \vspace{-6mm}
    \caption{Overview of the \model\ framework.}
    \label{fig:overall}
\end{figure*}

\section{Related Work}
\stitle{Urban Datasets and Benchmarks.}
Existing urban datasets and benchmarks target specific aspects of spatial understanding, such as navigation~\cite{chen2019touchdown}, urban scene understanding~\cite{hu2022sensaturban,cordts2015cityscapes}, urban perception~\cite{hou2024global,salesses2012place}, geolocation prediction~\cite{haas2024pigeon,wu2022im2city,hays2008im2gps}, and land-use analysis~\cite{wang2021loveda}. While effective for task-specific learning, these datasets lack explicit alignment between images, text, and urban spatial structure. CityBench~\cite{feng2025citybench} evaluates LLMs across diverse urban tasks, revealing limitations in spatial reasoning. UrbanLLaVA~\cite{feng2025urbanllava}, CityCube~\cite{xu2026citycube}, SpatialLab \cite{wasi2026spatialab}, and InternSpatial~\cite{deng2025internspatial} assess VLMs’ spatial reasoning ability through QA-style tasks largely confined to the camera’s observable field. DynamicVL~\cite{xuan2025dynamicvl} and URBANFEEL~\cite{he2025urbanfeel} focus on dynamic and perceptual city understanding, while GeoBenchX~\cite{krechetova2025geobenchx} and GeoChain~\cite{yerramilli2025geochain} evaluate multi-step geographic reasoning and expose LLM/VLMs’ weaknesses in spatial inference. Overall, existing urban datasets and benchmarks remain task-centric and underexplore the quality of learned spatial representations. In contrast, we propose a spatially grounded training dataset, together with a representation-centric benchmark that evaluates how well multimodal embeddings encode and align urban spatial knowledge across modalities.


\stitle{Multimodality Learning.}
Recent visual language models (VLMs) such as LLaVA and Qwen-VL achieve strong visual understanding via large-scale instruction tuning, while multimodal embedding models extend CLIP-style alignment toward more general representation spaces that accommodate diverse modalities. Approaches including GME\cite{zhang2024gme}, jina-embeddings-v4 \cite{gunther2025jina}, and VLM2Vec/VLM2Vec
-V2 \cite{jiang2024vlm2vec,meng2025vlm2vec} use contrastive training to learn transferable embeddings across images, text, and other multimodal content. Despite the progress, both VLMs and general-purpose multimodal embeddings predominantly encode visual, semantic, or layout-level patterns, while lacking explicit representations of geographically embedded structure such as connectivity and topology, which limits their applicability to urban tasks that require spatial understanding beyond local visual cues. The absence of datasets and benchmarks with explicit spatial structure also constrains systematic training and evaluation of such capabilities of VlMs.

\stitle{Graph-based Spatial Encoding for Urban Knowledge.}
A complementary line of work enhances LLM and VLM reasoning by grounding models in structured relational data such as knowledge graphs (KG) and text-attributed graphs. \cite{lee2023vista,peng2023kosmos,tandon2014webchild,ilievski2021cskg} demonstrated that visual–textual knowledge graphs enable relational grounding. GCR\cite{luo2410graph} has showed that graph-constrained reasoning improves faithfulness and interpretability in language-centric settings. \cite{hu2024geometric} introduced geometric KG embeddings that incorporate spatial cues for structured reasoning but was not designed for urban-scale multimodal alignment. Recent frameworks such as SpatialRGPT \cite{cheng2024spatialrgpt} and SpatialRAG \cite{yu2025spatial} leveraged structured spatial information for question answering and reasoning, yet treat spatial structure as auxiliary context rather than a representation modality aligned with visual perception. In parallel, Structure-CLIP \cite{huang2024structure} showed that graph-conditioned learning can enhance vision--language embeddings, though it focused on object-centric scene graphs rather than geographic structure. Prior GeoAI work applied graph encoders to model distance, direction, and topology \cite{klemmer2023positional, lee2024latent, chu2025geo2vec}. Collectively, these studies motivate the explicit integration of graph-structured spatial priors into VLMs, enabling spatially aware multimodal embeddings that generalize across diverse urban science tasks.

\section{\model}
In this section, we present \model\ (\modelS). As illustrated in Fig.~\ref{fig:overall}, the \modelS\ framework integrates spatially grounded training data (\dataset), a progressive two-stage learning strategy, and a unified benchmark for evaluating urban spatial representations (\benchmark). We introduce these components in turn.
\subsection{\dataset: Training Data Pipeline}\label{sec.dataset}
\subsubsection{Urban Spatial Graph Construction.}
We construct city-scale spatial graphs by integrating street-level imagery from Mapillary\footnote{\url{https://www.mapillary.com/}}
 with urban entities extracted from OpenStreetMap\footnote{\url{https://www.openstreetmap.org/}}, further augmented by open municipal datasets covering buildings, green space, historic districts, and transportation\footnote{\url{https://opendata.cityofnewyork.us/}, \url{https://data.gov.sg/}}. In this graph, nodes represent street-view images or urban entities, while edges encode spatial relations (Table~\ref{tab:spatial-graph-schema}), capturing accessibility and connectivity to support spatial reasoning. This spatial graph is naturally scalable, allowing new entities to be incorporated via local spatial relations without global reconstruction.

Based on the spatial graphs, we derive \emph{spatial reasoning paths} (SRPs) for Stage~1 training and \emph{spatial context captions} (SCCs) for Stage~2 graph-conditioned training, as follows.
\begin{table}[t]
\caption{Urban spatial graph schema.}
\label{tab:spatial-graph-schema}
\vspace{-0.5em}
\centering
\footnotesize
\addtolength{\tabcolsep}{4pt}
\renewcommand{\arraystretch}{1.0}
\begin{tabular}{ccl}
\toprule
\textbf{Category} & \textbf{Type} & \textbf{Description} \\
\midrule
\multirow{6}{*}{Node} 
& Street-level viewpoint & Street-view image location \\
& Road & Road \\
& Intersection & Junction of roads \\
& POI & Point of Interests \\
& AOI & Area of Interests \\
& Transit Facility & Metro station, bus stop, etc. \\
\midrule
\multirow{5}{*}{Edge} 
& on\_same\_street & Two POIs on the same road \\
& crossing & Junction points of roads \\
& nearest & POI's nearest road \\
& near & A road is near to a POI \\
& bounds & A road bounds AOI \\
& intersects & A road intersects AOI \\
\bottomrule
\end{tabular}
\vspace{-0.8em}
\end{table}

\subsubsection{Spatial Reasoning Paths.}
A key challenge in spatially aware multimodal learning is bridging structured urban graphs with the language-centric reasoning of VLMs. To address this, we introduce Spatial Reasoning Paths (SRPs), which extend reasoning paths in RoG \cite{luo2023reasoning} and GCR \cite{luo2410graph} by revealing relative distance and direction and organizing graph triples into a sequence of waypoints for navigation-based spatial alignment. This design is motivated by cognitive theories showing that human form mental maps through relational and hierarchical structures centered on nodes, paths, and landmarks rather than isolated locations \cite{lynch1960image,siegel1975development,stevens1978distortions,mcnamara1986mental}.

Specifically, an SRP is an ordered sequence of relational triples $(e_s, r, e_t)$, capturing semantic and topological relations (e.g.,
\texttt{nearest}, \texttt{crossing}), interleaved with transition tuples $(d, \theta)$ that encode geodesic distance and relative direction between consecutive entities based on geographic coordinates and network geometry. An example SRP is shown below:
\vspace{0.5em}

\noindent
\fbox{
\footnotesize
\begin{tabular}{@{}l@{}}
$\big(\{\text{imageID}\},\ \texttt{nearest},\ \text{Robert F. Kennedy Bridge}\big)
(10.6\,\text{m},\,52^{\circ}\text{NE})$\\
$\xrightarrow{(789.0\,\text{m},\,143^{\circ}\text{SE})}
\big(\text{Robert F. Kennedy Bridge},\ \texttt{crossing},\ \text{21st St}\big)$\\
$\xrightarrow{(310.1\,\text{m},\,223^{\circ}\text{SW})}
\big(\text{21st St},\ \texttt{near},\ \text{NY Ctr.~for Rehab \& Nursing}\big)$
$(28.5\,\text{m},\,128^{\circ}\text{SE}).$
\end{tabular}
\normalsize
}

\begin{figure}[t]
    \centering
    \includegraphics[width=\columnwidth]{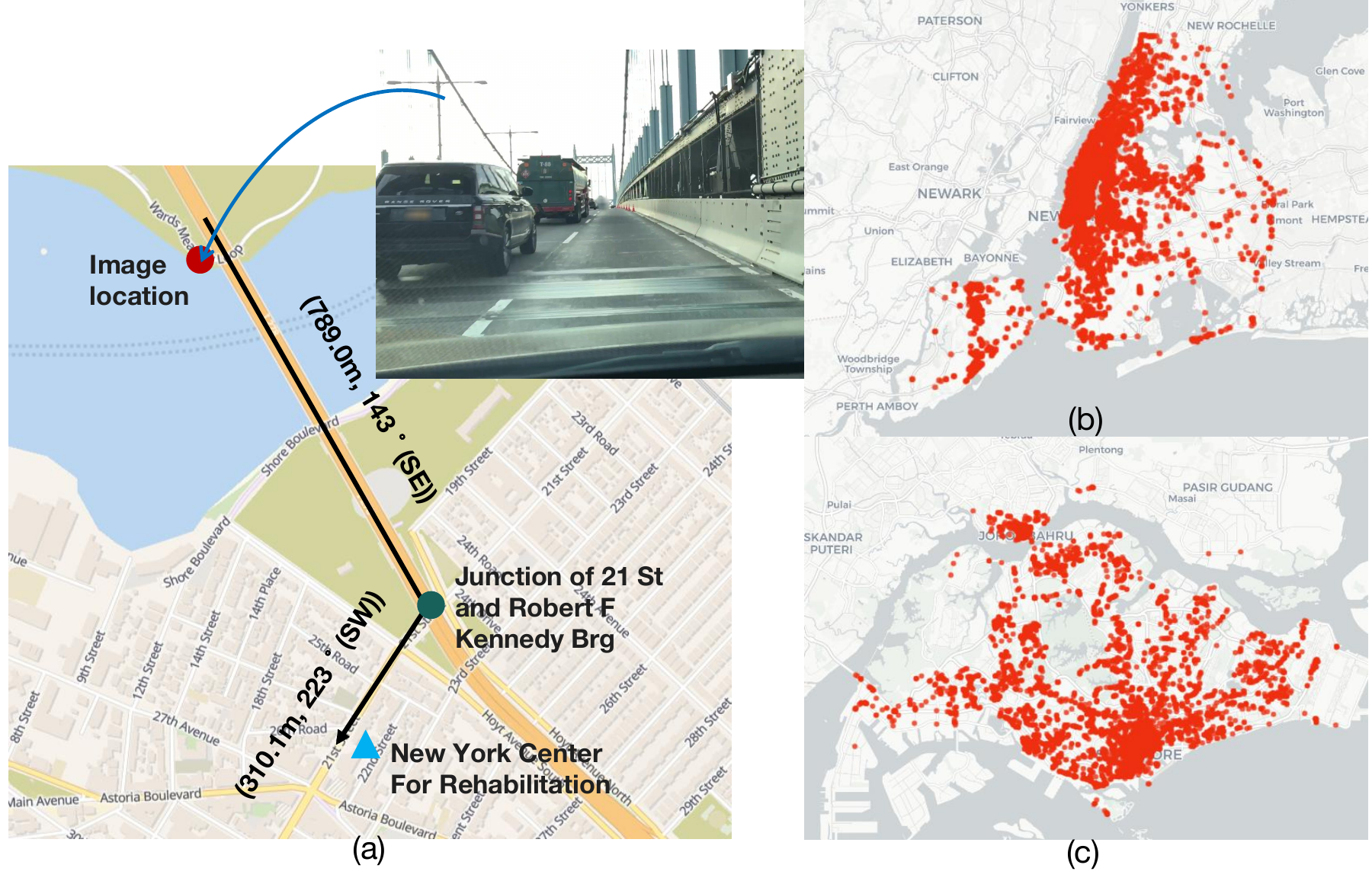}
    \vspace{-6mm}
    \caption{
    Overview of spatial context and data coverage:
    \textbf{(a)} Map illustration of an SRP;
    \textbf{(b), (c)} Street-view image distributions in New York and Singapore.
    }
    \vspace{-0.8em}
    \label{fig:maps}
\end{figure}

\vspace{0.5em}
\noindent From the image location, the path links to the Robert F. Kennedy Bridge (10.6 m northeast), follows the road southeast for 789 m to the bridge--21st Street crossing, and reaches NY Center for Rehab \& Nursing, 28.5 m southeast of 21st Street, as illustrated in Fig.~\ref{fig:maps}(a).

In implementation, we generate SRPs via S2-based partitioning\footnote{S2 is a spherical geometry library for representing geographic data and building spatial indexes (\url{https://s2geometry.io/}).}
 and localized graph traversal. See Appendix~\ref{app:srp} for SRP generation details.
%
We also include brief instructional text accompanying each SRP to indicate how the path should be interpreted from a user perspective (e.g., reachability, facing direction, and viewpoint), guiding the model to ground the path in both the image and spatial graph. See examples in Appendix~\ref{app:data_samples}.

\subsubsection{Spatial Context Captions.}
While SRPs focus on individual paths originating from the image location, they lack a holistic representation of the surrounding neighborhood.
To capture such a holistic view, thereby better aligning street-level images with their spatial graphs, we generate Spatial Context Captions (SCCs) that integrate visual cues with graph-derived semantic and relational information. For each image, a localized spatial subgraph is summarized as structured context and, together with the image, provided to an LLM (GPT-4.1 and Qwen-VL-Max-Latest) to generate a spatially grounded natural-language caption\footnote{The captioning model is instructed to rely strictly on the provided subgraph description and not introduce external or fabricated spatial information.} (see details and examples in Appendix~\ref{app:data_samples}, \ref{app:scc}, \ref{app:sub_des}). These captions jointly capture what is visible in the image and how the location is situated within its urban neighborhood.
\subsubsection{Data Statistics.}
Our dataset is built on city-scale spatial graphs from New York, Singapore, Beijing, and Paris, with New York and Singapore used for training and Beijing and Paris reserved for cross-city evaluation. The spatial graphs for New York and Singapore contain 487K and 338K nodes and 7.3M and 333K edges, respectively. From the two training cities, we construct 17,736 SRPs (average 7.27 hops) and generate 8,039 SCCs, comprising 92,612 entity mentions over 20,838 unique entities. These graph-derived signals are used to construct supervision at different stages of training: \emph{Stage~1} includes 44,356 image--text samples (SRPs and image-level captions), while \emph{Stage~2} consists of 25,775 graph-conditioned samples (SRPs and SCCs with graphs). The spatial distribution of street-view images in the training cities is shown in Fig.~\ref{fig:maps}(b,c). Overall, our dataset pipeline is scalable, extensible, and reproducible. 
Next, we introduce the two-stage progressive training strategy using our data. 

\subsection{Two-Stage Training}\label{sec.model}
\vspace{0.5em}
While SRPs and SCCs provide spatially grounded supervision at the data level, effectively incorporating structured spatial knowledge into VLM embeddings poses additional challenges at the model and optimization level. First, urban spatial knowledge is heterogeneous and structured, involving connectivity, topology and neighborhood context that are not explicitly encoded in standard image–text pairs. Second, directly injecting such knowledge through task-level supervision often leads to representations that do not generalize well across retrieval, ranking, and reasoning settings. Finally, na\"ively mixing visual, textual, and graph-structured signals during training can lead to unstable optimization and under-utilization of spatial structure. To address these challenges, we adopt a two-stage training strategy tailored to spatial VLM embedding learning (Fig.~\ref{fig:overall}).

\subsubsection{Stage 1: Instruction-Guided Image--Text Alignment.}
In the first stage, we perform \emph{instruction-guided contrastive embedding learning} to align images with spatially enriched textual descriptions, including \emph{SRPs} and \emph{image-level captions}. Beyond generic image--text alignment, this stage injects structured spatial knowledge into the visual--language embedding space by grounding images in local spatial context expressed through language, following the contrastive paradigm of VLM2Vec \cite{jiang2024vlm2vec}.

Let $(q, t^{+})$ denote a positive query--target pair, where the query $q$ is a street-level image and the target $t^{+}$ is a textual description. 

\stitle{Instruction construction.}
Given an image query $q_i$, we construct an instruction-augmented query $q_{\text{inst}}$ as:
\begin{equation}
q_{\text{inst}} = [\text{IMAGE\_TOKEN}] \ \text{Instruct: } \{\tau\},
\label{eq:stage1_inst}
\end{equation}
where $\tau$ is a fixed, task-agnostic instruction selected according to the target type. Specifically, we use two instruction templates:

\vspace{2mm}
\noindent
\fbox{\small\begin{minipage}{0.97\linewidth}%
\vspace{-2mm}%
\begin{align}
\tau_{\text{path}}:\ &\text{``Describe whether the \{destination\} is reachable from} \nonumber\\[-0.5mm]
&\text{this viewpoint.''} \nonumber\\
\tau_{\text{cap}}:\ &\text{``Provide a detailed description of the image content.''}
\nonumber
\end{align}
\end{minipage}}

\vspace{2mm}
\noindent These instructions guide the model to interpret the image either in terms of \emph{navigability and spatial relations} or \emph{visual semantics}.

\stitle{Contrastive objective.}
We optimize the embeddings using the standard InfoNCE loss over in-batch negatives:
\begin{equation}
\mathcal{L}_{\text{Stage1}} =
-\log
\frac{
\phi\!\left(\mathbf{h}_{q_{\text{inst}}}, \mathbf{h}_{t^{+}}\right)
}{
\phi\!\left(\mathbf{h}_{q_{\text{inst}}}, \mathbf{h}_{t^{+}}\right)
+
\sum_{t^{-} \in \mathcal{N}}
\phi\!\left(\mathbf{h}_{q_{\text{inst}}}, \mathbf{h}_{t^{-}}\right)
},
\label{eq:stage1_infonce}
\end{equation}
where $\mathcal{N}$ denotes in-batch negative text samples and $\phi(\cdot,\cdot)$ is a similarity function (cosine similarity with temperature scaling).

\subsubsection{Stage 2: Graph-Conditioned Spatial Embedding.}
In the second stage, we introduce a \emph{graph encoder} that explicitly conditions street-view image embeddings on their associated local spatial graphs, enabling alignment with structured urban spatial context.
We again consider a positive pair $(q, t^{+})$. In Stage~2, the query is \emph{multimodal}, consisting of a street-level image and its associated spatial graph, while the target remains text-based (SRPs or SCCs).

\stitle{Instruction construction with graph tokens.}
Given an image $q_i$ and its corresponding spatial subgraph $q_g$, we construct an instruction-augmented multimodal query:
\begin{equation}
q_{\text{inst}} = [\text{GRAPH\_TOKEN}][\text{IMAGE\_TOKEN}] \ \text{Instruct: } \{\tau\},
\end{equation}
where $\tau$ is a fixed instruction selected according to the target type. We use two instruction templates in Stage~2:

\vspace{2mm}
\noindent
\fbox{\small\begin{minipage}{0.97\linewidth}%
\vspace{-2mm}%
\begin{align}
\tau_{\text{path}}:\ &\text{``Refer to the image and spatial graph, describe the pedestrian} \nonumber\\[-0.5mm]
&\text{navigation context and spatial paths from this viewpoint.''} \nonumber\\
\tau_{\text{ctx}}:\ &\text{``Use the image and graph together to describe the scene and} \nonumber\\[-0.5mm]
&\text{its spatial context.''}\nonumber
\end{align}
\end{minipage}}


\vspace{2mm}
\stitle{Progressive injection.}
We apply the same InfoNCE loss for Stage 2, as introduced in Eq.~\eqref{eq:stage1_infonce}.
To preserve the visual--language alignment learned in Stage~1, we train the graph encoder with a higher learning rate while updating VLM parameters conservatively, allowing spatial structure to be injected without disrupting visual grounding. This staged design decouples semantic alignment from spatial regularization and supports flexible inference with or without graph input, yielding transferable multimodal embeddings. We next introduce the graph encoder used in Stage~2.

\subsubsection{Graph Encoder (for Stage 2).}
The graph encoder is composed of a node encoder and an edge encoder, designed to encode localized urban subgraphs surrounding each street image. The graph encoder transforms spatial entities and their pairwise spatial relations into embeddings that can be fused with image representations for graph-conditioned contrastive learning.

\stitle{Node encoder.}
Each node is associated with textual attributes (e.g., name, category, address) and geographic coordinates. Textual attributes are encoded using the shared language embedding layer of the VLM. To encode spatial position, we apply a multi-frequency sinusoidal spatial positional encoding over longitude–latitude coordinates, following the design principles of PE-GNN\cite{klemmer2023positional}. The textual and spatial embeddings are concatenated and projected to obtain the final node embedding.

\stitle{Edge encoder.}
Edges encode explicit geodesic and spatial relationships between nodes, capturing how urban entities are positioned and connected in physical space. For each edge, we construct three complementary spatial features:
(1) relative distance, computed as the log-transformed haversine distance;
(2) relative direction, represented by sine and cosine of the bearing angle;
(3) relative displacement, given by the coordinate offset between nodes.
These features are jointly embedded through a learnable edge encoder.
 
\stitle{Graph message passing.}
We adopt GATv2\cite{brody2021attentive} conditioned on learned edge features encoding spatial relations, enabling multi-layer propagation of spatial context with residual connections and layer normalization.

\begin{table}[t]
\centering
\caption{Overview of \benchmark\ tasks and dataset statistics.}
\vspace{-0.5em}
\label{tab:benchmark}
\setlength{\tabcolsep}{3pt}
\renewcommand{\arraystretch}{1.0}
\begin{threeparttable}
\resizebox{\columnwidth}{!}{
\begin{tabular}{@{}lccc|cccc@{}}
\toprule
\textbf{Task} & \textbf{Query} & \textbf{Target} & \textbf{\#Cand.} & \textbf{NY} & \textbf{SG} & \textbf{BJ} & \textbf{PA} \\
\midrule
\textbf{Geolocation} & I & T & 20 & 260 & 260 & 260 & 260 \\
\midrule
\multicolumn{8}{l}{\textbf{Urban Perception}} \\
Safe, Wealthy, Lively, Depressing, Boring, Beautiful 
& I & T & 20 & 389 & 389 & 389 & 389 \\
\midrule
\textbf{Image Retrieval} & T / T+G & I & 20 & 700 & 700 & 700 & 700 \\
\midrule
\multicolumn{8}{@{}l}{\textbf{Spatial Grounding}} \\
Nearest Street & I / I+G & T & 20 & 600 & 600 & 600 & 600 \\
Nearest POI & I / I+G & T & 20 & 260 & 260 & 260 & 260 \\
Distance & I / I+G & T & 20 & 260 & 260 & 260 & 260 \\
Distance-Direction & I / I+G & T & 20 & 260 & 260 & 260 & 260 \\
\bottomrule
\end{tabular}
}
\begin{tablenotes}[flushleft]
\scriptsize
\item \textbf{I}/\textbf{T}/\textbf{G} denotes image/text/spatial graph queries; \textbf{I+G}/\textbf{T+G} indicates multimodal combinations. \\
\end{tablenotes}
\end{threeparttable}
\vspace{-2mm}
\end{table}

\subsection{UGBench: Evaluating Urban Tasks}
\label{sec.benchmark}
We introduce the \textbf{\benchmark} benchmark, which evaluates spatially informed VLM embeddings across four tasks: \emph{geolocation ranking}, \emph{image retrieval}, \emph{urban perception}, and \emph{spatial grounding}. An overview of the benchmark is provided in Table~\ref{tab:benchmark}.

All tasks in \benchmark\ are formulated as \emph{embedding-based ranking problems}, following VLM2Vec~\cite{jiang2024vlm2vec}. Each instance provides an instruction and a query (text, image, or image with graph context), and both queries and candidates are embedded into a shared space using the last-layer hidden state of the final token. Candidates are then ranked by cosine similarity to the query embedding. Detailed descriptions of each task follow.

\subsubsection{Geolocation Ranking.}
Unlike conventional geolocation benchmarks that predict coordinates or coarse regions, our geolocation ranking task uses \emph{fine-grained textual descriptions of local spatial context} as labels, capturing hierarchical and relational cues such as street names, intersections, nearby landmarks, and neighborhoods anchored to precise coordinates.

\vspace{1mm}
\noindent
\fbox{%
\begin{minipage}{0.97\linewidth}\small
\textbf{Example.} \emph{Query:} ``\{image\} Where is this street view image taken? Select the correct location.'' 
\emph{Candidates:} ``on Union St, near the intersection of 5 Ave and Union St, in Park Slope, Brooklyn, ...''

\noindent\textbf{What it tests.} This task evaluates fine-grained, hierarchical, and relational location understanding beyond coordinate or region prediction.
\end{minipage}}

\subsubsection{Image Retrieval}
The task is to retrieve relevant street view images given a textual query augmented with spatial graph context.

\vspace{1mm}
\noindent
\fbox{%
\begin{minipage}{0.97\linewidth}\small
\textbf{Example.} \emph{Queries:} ``\{graph\} Which street view image references a journey ending at 248, Central Park West?''; ``\{graph\} Using the spatial graph provided, find the image whose spatial context matches: This image is situated within a dense urban network centered around Queens Boulevard...''
\emph{Candidates:} 21 Mapillary images from the same city corpus.

\noindent\textbf{What it tests.} This task evaluates cross-modal alignment between textual semantics, structured spatial context, and visual content, and whether explicit spatial structure improves retrieval beyond text alone.
\end{minipage}}

\subsubsection{Urban Perception}
This benchmark extends \emph{Global Streetscapes} \cite{hou2024global} by augmenting each street-view image with its associated \emph{local spatial graph}. Human-annotated perception scores (safety, wealthiness, liveliness, beauty, boredom, and depressiveness) are discretized into fine-grained textual categories, and the task is to rank these labels given an image and its spatial graph. For evaluation, we randomly sample 389 held-out instances from New York and Singapore after excluding training locations.

\vspace{1mm}
\noindent
\fbox{%
\begin{minipage}{0.97\linewidth}\small
\noindent\textbf{Example.} \emph{Query:} ``\{graph\}\{image\} Given the spatial graph, what is the perception of wealthiness for this image location?'' 
\emph{Candidates:} ``Very wealthy, Wealthy Score: 9.0,'' ``Moderately wealthy, Wealthy Score: 6.5, ...''

\noindent\textbf{What it tests.} This task evaluates whether embeddings capture higher-level urban semantics that are not fully determined by local visual appearance and may benefit from broader spatial context.
\end{minipage}}

\subsubsection{Spatial Grounding}
Compared to previous spatial reasoning benchmarks \cite{deng2025internspatial,feng2025citygpt,feng2025urbanllava,xu2026citycube}, our spatial grounding task does not supervise explicit answer generation, but evaluates whether spatial structure underlying \emph{daily experience} is internalized in the embedding space to support retrieval and transfer across urban tasks. Queries and candidates are derived from a \emph{local spatial graph anchored at a street-view image location} and span three types: (1) proximity queries, (2) distance-sensitive queries based on relative distance, and (3) compositional distance-direction queries involving directionality and multi-hop relations.

\vspace{1mm}
\noindent
\fbox{%
\begin{minipage}{0.97\linewidth}\small
\noindent\textbf{Examples.}
\begin{itemize}[leftmargin=*]
    \item \emph{Nearest street:} ``\{graph\}\{image\} Using the provided spatial graph, what street is this location on or nearest to?''
    \item \emph{Nearest POI:} ``\{graph\}\{image\} Using the provided spatial graph, which POI is nearest to the current image location?''
    \item \emph{Distance:} ``\{graph\}\{image\} How far is the nearest church from the closest commercial?''
    \item \emph{Distance-direction:} ``\{graph\}\{image\} Refer to the spatial graph, if you walk approximately 200 meters south from here, what landmark will you encounter?''
\end{itemize}

\noindent\textbf{What it tests.}
This benchmark evaluates whether spatial reasoning concepts, such as proximity, distance, directionality, and multi-hop relations, are implicitly reflected in the embedding space to enable spatially informed retrieval.
\end{minipage}}

\begin{table*}[t]
\centering
\caption{Results on geolocation ranking, image retrieval, and urban perception. Urban perception is averaged over 6 attributes listed in Table~\ref{tab:benchmark}; results for each individual attribute are reported in Appendix~\ref{app:perception_tasks}. H@5: Hit@5, N@5: NDCG@5, in percent.}
\vspace{-0.5em}
\label{tab:geo_retrieval}
\footnotesize
\setlength{\tabcolsep}{2.5pt}
\renewcommand{\arraystretch}{1.05}
\resizebox{1\textwidth}{!}{%
\begin{tabular}{l cc|cc|cc|cc|cc|cc|cc|cc|cc|cc|cc|cc}
\toprule
\multicolumn{1}{c}{\textbf{Model}}
& \multicolumn{8}{c}{\textbf{Geolocation Ranking}} 
& \multicolumn{8}{c}{\textbf{Image Retrieval}}
& \multicolumn{8}{c}{\textbf{Urban Perception (Avg.)}} \\
\cmidrule(lr){2-9} \cmidrule(lr){10-17} \cmidrule(lr){18-25}

& \multicolumn{2}{c}{NY} & \multicolumn{2}{c}{SG} & \multicolumn{2}{c}{BJ} & \multicolumn{2}{c}{PA}
& \multicolumn{2}{c}{NY} & \multicolumn{2}{c}{SG} & \multicolumn{2}{c}{BJ} & \multicolumn{2}{c}{PA}
& \multicolumn{2}{c}{NY} & \multicolumn{2}{c}{SG} & \multicolumn{2}{c}{BJ} & \multicolumn{2}{c}{PA} \\
\cmidrule(lr){2-3} \cmidrule(lr){4-5} \cmidrule(lr){6-7} \cmidrule(lr){8-9}
\cmidrule(lr){10-11} \cmidrule(lr){12-13} \cmidrule(lr){14-15} \cmidrule(lr){16-17}
\cmidrule(lr){18-19} \cmidrule(lr){20-21} \cmidrule(lr){22-23} \cmidrule(lr){24-25}

& H@5 & N@5 & H@5 & N@5 & H@5 & N@5 & H@5 & N@5
& H@5 & N@5 & H@5 & N@5 & H@5 & N@5 & H@5 & N@5
& H@5 & N@5 & H@5 & N@5 & H@5 & N@5 & H@5 & N@5 \\
\midrule

Qwen2-VL-2B
& 29.23 & 14.86 & 34.23 & 18.26 & 44.62 & 24.50 & 45.00 & 24.49
& 32.00 & 19.47 & 29.29 & 17.77 & 34.14 & 20.77 & 30.14 & 17.38
& 22.75 & 13.09 & 21.20 & 12.51 & 23.53 & 13.31 & 24.08 & 14.85 \\
\textbf{\modelS}
& \textbf{44.23} & \textbf{30.98} & \textbf{48.85} & \textbf{36.04} & \textbf{71.92} & \textbf{64.18} & \textbf{61.15} & \textbf{42.80}
& \textbf{55.57} & \textbf{39.36} & \textbf{62.23} & \textbf{45.72} & \textbf{62.57} & \textbf{40.61} & \textbf{49.07} & \textbf{33.42}
& \textbf{31.47} & \textbf{18.60} & \textbf{26.98} & \textbf{16.32} & \textbf{23.90} & \textbf{13.53} & \textbf{26.40} & \textbf{16.08} \\
\addlinespace[0.05ex]
\midrule
\addlinespace[0.05ex]

Qwen2-VL-7B
& 27.31 & 17.16 & 35.00 & 22.93 & 59.62 & 50.66 & 30.00 & 18.37
& 31.57 & 18.69 & 28.29 & 16.98 & 36.86 & 22.56 & 30.00 & 18.42
& 16.80 & 8.45 & 24.33 & 13.60 & \textbf{27.38} & \textbf{17.61} & 22.45 & 11.50 \\
\textbf{\modelS}
& \textbf{55.00} & \textbf{39.30} & \textbf{49.62} & \textbf{35.26} & \textbf{65.77} & \textbf{54.93} & \textbf{48.85} & \textbf{31.24}
& \textbf{65.43} & \textbf{47.32} & \textbf{70.10} & \textbf{52.44} & \textbf{56.29} & \textbf{39.06} & \textbf{49.07} & \textbf{34.12}
& \textbf{31.96} & \textbf{18.02} & \textbf{29.52} & \textbf{17.47} & 25.75 & 14.51 & \textbf{27.89} & \textbf{16.22} \\
\addlinespace[0.05ex]
\midrule
\addlinespace[0.05ex]

Qwen2.5-VL-3B
& 19.23 & 11.76 & 30.00 & 17.85 & 20.77 & 13.16 & 0.38 & 0.19
& 26.71 & 15.86 & 26.00 & 15.76 & 27.43 & 16.73 & 24.29 & 15.29
& 24.48 & \textbf{14.95} & 20.57 & 12.20 & 16.82 & 8.81 & 20.61 & 12.26 \\
\textbf{\modelS}
& \textbf{37.69} & \textbf{23.93} & \textbf{36.92} & \textbf{26.83} & \textbf{56.92} & \textbf{49.15} & \textbf{40.77} & \textbf{28.84}
& \textbf{47.43} & \textbf{31.12} & \textbf{45.06} & \textbf{30.33} & \textbf{44.29} & \textbf{29.50} & \textbf{36.05} & \textbf{22.68}
& \textbf{25.48} & 14.75 & \textbf{26.18} & \textbf{15.59} & \textbf{21.42} & \textbf{13.63} & \textbf{26.48} & \textbf{15.44} \\
\addlinespace[0.05ex]
\midrule
\addlinespace[0.05ex]

Qwen2.5-VL-7B
& 32.69 & 20.24 & 33.85 & 21.43 & 48.08 & 36.01 & 39.62 & 26.83
& 33.71 & 21.21 & 26.57 & 15.97 & 27.14 & 15.98 & 31.71 & 18.98
& 20.44 & 11.18 & 26.18 & 15.00 & 27.08 & 16.75 & 23.82 & 13.32 \\
\textbf{\modelS}
& \textbf{58.85} & \textbf{47.94} & \textbf{64.23} & \textbf{54.65} & \textbf{71.92} & \textbf{63.38} & \textbf{65.77} & \textbf{52.88}
& \textbf{58.86} & \textbf{43.24} & \textbf{71.10} & \textbf{52.22} & \textbf{57.43} & \textbf{40.44} & \textbf{57.22} & \textbf{41.50}
& \textbf{41.69} & \textbf{27.12} & \textbf{35.35} & \textbf{23.68} & \textbf{34.17} & \textbf{22.71} & \textbf{40.59} & \textbf{26.81} \\
\addlinespace[0.05ex]
\midrule
\addlinespace[0.05ex]

GME-2B
& 29.23 & 14.86 & 34.23 & 18.26 & 44.62 & 24.50 & 45.00 & 24.49
& 42.43 & 22.78 & 47.29 & 25.60 & 53.14 & 28.70 & 44.00 & 23.56
& \textbf{28.03} & \textbf{14.26} & 28.24 & 13.85 & \textbf{30.04} & \textbf{14.63} & \textbf{26.57} & \textbf{12.94} \\
\textbf{\modelS}
& \textbf{41.15} & \textbf{28.68} & \textbf{46.15} & \textbf{33.15} & \textbf{46.54} & \textbf{32.20} & \textbf{54.62} & \textbf{37.81}
& \textbf{55.43} & \textbf{41.16} & \textbf{60.37} & \textbf{44.10} & \textbf{59.14} & \textbf{40.80} & \textbf{51.65} & \textbf{36.06}
& 25.04 & 13.12 & \textbf{30.24} & \textbf{17.52} & 20.91 & 12.60 & 19.19 & 10.76 \\
\addlinespace[0.05ex]
\midrule
\addlinespace[0.05ex]

VLM2Vec-2B
& 30.77 & 18.54 & 34.23 & 22.06 & \textbf{43.85} & \textbf{33.82} & \textbf{56.15} & \textbf{34.97}
& 52.57 & 36.30 & 56.43 & 42.17 & \textbf{59.29} & \textbf{43.72} & \textbf{54.00} & \textbf{37.50}
& 18.27 & 9.01 & 18.59 & 8.76 & 18.92 & 8.86 & 19.02 & 9.21 \\
\textbf{\modelS}
& \textbf{48.85} & \textbf{31.64} & \textbf{47.69} & \textbf{33.79} & 39.23 & 29.08 & 50.00 & 34.50
& \textbf{54.57} & \textbf{38.67} & \textbf{63.52} & \textbf{45.59} & 58.29 & 38.52 & 48.93 & 33.01
& \textbf{31.02} & \textbf{17.56} & \textbf{25.24} & \textbf{14.88} & \textbf{24.43} & \textbf{13.66} & \textbf{25.07} & \textbf{14.21} \\
\addlinespace[0.05ex]
\midrule
\addlinespace[0.05ex]

LLaVA1.6
& 28.85 & 17.73 & 26.15 & 14.80 & 22.69 & 14.77 & 20.00 & 11.48
& 29.71 & 17.60 & 29.29 & 18.22 & \textbf{32.29} & \textbf{18.74} & 28.00 & 16.98
& 29.01 & 17.50 & 23.39 & 13.85 & \textbf{24.64} & \textbf{14.75} & \textbf{24.55} & \textbf{13.75} \\
\textbf{\modelS}
& \best{53.46} & \best{35.13} & \best{55.38} & \best{38.06} & \best{48.08} & \best{37.64} & \best{56.92} & \best{39.54}
& \best{41.50} & \best{28.64} & \best{39.41} & \best{27.59} & 31.50 & 18.27 & \best{46.92} & \best{30.96}
& \best{35.54} & \best{21.19} & \best{24.52} & \best{14.22} & 23.35 & 13.91 & 24.11 & 13.75 \\
\addlinespace[0.05ex]
\midrule
\addlinespace[0.05ex]

Phi3-V
& 24.62 & 13.53 & 28.85 & 17.30 & 38.46 & 28.67 & 18.85 & 12.16
& 36.29 & 21.04 & 36.14 & 23.04 & 31.00 & 19.65 & 31.00 & 18.57
& 27.16 & 15.04 & 31.79 & 18.21 & 24.12 & 13.74 & 25.96 & 14.73 \\
\textbf{\modelS}
& \best{39.23} & \best{23.70} & \best{43.85} & \best{30.96} & \best{51.92} & \best{43.14} & \best{58.46} & \best{41.36}
& \best{45.43} & \best{29.73} & \best{47.93} & \best{30.24} & \best{48.43} & \best{31.11} & \best{41.34} & \best{26.26}
& \best{28.55} & \best{16.17} & \best{23.31} & \best{13.46} & \best{27.68} & \best{16.66} & \best{27.68} & \best{16.07} \\

\bottomrule
\end{tabular}}
\end{table*}

\begin{table*}[t]
\centering
\caption{Spatial grounding results across query types and cities. Results of \emph{Nearest Location} is reported in Appendix~\ref{app:nearest_street}.}
\vspace{-0.5em}
\label{tab:spatial_grounding_all}
\footnotesize
\setlength{\tabcolsep}{2.5pt}
\renewcommand{\arraystretch}{1.05}
\resizebox{1\textwidth}{!}{
\begin{tabular}{l cc|cc|cc|cc|cc|cc|cc|cc|cc|cc|cc|cc}
\toprule
\multicolumn{1}{c}{\textbf{Model}}
& \multicolumn{8}{c}{\textbf{Nearest POI}}
& \multicolumn{8}{c}{\textbf{Distance}}
& \multicolumn{8}{c}{\textbf{Distance--Direction}} \\
\cmidrule(lr){2-9}\cmidrule(lr){10-17}\cmidrule(lr){18-25}

& \multicolumn{2}{c}{NY} & \multicolumn{2}{c}{SG} & \multicolumn{2}{c}{BJ} & \multicolumn{2}{c}{PA}
& \multicolumn{2}{c}{NY} & \multicolumn{2}{c}{SG} & \multicolumn{2}{c}{BJ} & \multicolumn{2}{c}{PA}
& \multicolumn{2}{c}{NY} & \multicolumn{2}{c}{SG} & \multicolumn{2}{c}{BJ} & \multicolumn{2}{c}{PA} \\
\cmidrule(lr){2-3}\cmidrule(lr){4-5}\cmidrule(lr){6-7}\cmidrule(lr){8-9}
\cmidrule(lr){10-11}\cmidrule(lr){12-13}\cmidrule(lr){14-15}\cmidrule(lr){16-17}
\cmidrule(lr){18-19}\cmidrule(lr){20-21}\cmidrule(lr){22-23}\cmidrule(lr){24-25}

& H@5 & N@5 & H@5 & N@5 & H@5 & N@5 & H@5 & N@5
& H@5 & N@5 & H@5 & N@5 & H@5 & N@5 & H@5 & N@5
& H@5 & N@5 & H@5 & N@5 & H@5 & N@5 & H@5 & N@5 \\
\midrule

Qwen2-VL-2B
& 39.44 & 29.72 & 45.38 & 42.44 & 66.92 & 58.65 & 26.92 & 19.29
& 12.31 & 11.01 & 18.85 & 16.90 & 27.69 & 26.09 & 23.46 & 22.06
& 26.09 & 24.99 & 23.46 & 22.06 & 40.00 & 39.40 & 36.54 & 35.15 \\
\textbf{\modelS}
& \best{45.07} & \best{37.18} & \best{60.00} & \best{55.77} & \best{72.31} & \best{66.75} & \best{30.77} & \best{20.08}
& \best{21.92} & \best{21.70} & \best{29.23} & \best{28.23} & \best{31.92} & \best{30.45} & \best{25.00} & \best{24.76}
& \best{33.04} & \best{32.04} & \best{26.54} & \best{26.30} & \best{41.92} & \best{39.80} & \best{45.00} & \best{40.40} \\
\addlinespace[0.05ex]
\midrule
\addlinespace[0.05ex]

Qwen2-VL-7B
& 36.15 & 29.19 & 38.46 & 36.41 & 58.08 & 51.02 & 23.08 & 14.90
& 23.85 & 22.85 & 14.23 & 13.03 & \textbf{30.38} & \textbf{29.04} & 17.69 & 16.09
& 18.26 & 18.01 & 24.23 & 22.00 & 28.85 & 27.05 & 34.23 & 33.90 \\
\textbf{\modelS}
& \best{43.66} & \best{36.10} & \best{62.31} & \best{59.65} & \best{75.38} & \best{68.76} & \best{29.23} & \best{19.97}
& \best{23.88} & \best{22.85} & \best{19.23} & \best{18.13} & 21.92 & 20.81 & \best{25.77} & \best{24.07}
& \best{33.91} & \best{32.01} & \best{34.23} & \best{33.03} & \best{51.65} & \best{50.00} & \best{40.84} & \best{40.00} \\
\addlinespace[0.05ex]
\midrule
\addlinespace[0.05ex]

Qwen2.5-VL-3B
& 36.15 & 26.01 & \textbf{50.00} & \textbf{47.81} & \textbf{76.15} & \textbf{63.14} & 25.77 & 17.08
& 26.92 & 25.02 & \textbf{30.77} & \textbf{28.89} & \textbf{44.62} & \textbf{42.79} & 25.00 & 22.76
& 28.70 & 27.70 & 26.92 & 25.12 & 25.77 & 25.17 & 29.62 & 28.21 \\
\textbf{\modelS}
& \best{37.09} & \best{29.72} & 41.54 & 39.70 & 65.38 & 56.67 & \best{29.62} & \best{21.79}
& \best{28.08} & \best{27.20} & 24.62 & 23.02 & 29.62 & 28.54 & \best{32.31} & \best{31.88}
& \best{29.57} & \best{28.30} & \best{31.15} & \best{30.03} & \best{38.85} & \best{37.33} & \best{38.08} & \best{36.89} \\
\addlinespace[0.05ex]
\midrule
\addlinespace[0.05ex]

Qwen2.5-VL-7B
& 39.91 & 30.49 & 58.85 & 54.93 & \textbf{76.15} & \textbf{65.19} & 27.10 & 18.54
& 16.15 & 14.21 & 9.23 & 7.03 & 12.69 & 10.15 & 22.31 & 20.17
& 26.09 & 24.11 & 27.31 & 27.31 & 32.69 & 30.23 & 35.77 & 34.17 \\
\textbf{\modelS}
& \best{55.87} & \best{48.60} & \best{70.77} & \best{68.58} & 70.62 & 59.89 & \best{28.08} & \best{19.12}
& \best{22.55} & \best{20.90} & \best{34.62} & \best{33.62} & \best{36.15} & \best{35.15} & \best{48.85} & \best{46.46}
& \best{44.35} & \best{43.30} & \best{29.23} & \best{28.03} & \best{58.95} & \best{57.71} & 50.38 & 49.80 \\
\addlinespace[0.05ex]
\midrule
\addlinespace[0.05ex]

GME-2B
& 35.21 & 20.608 & 58.85 & 36.41 & 64.62 & 37.54 & 23.08 & 12.20
& 25.00 & 15.77 & 17.31 & 10.92 & \textbf{21.92} & 13.83 & 13.46 & 8.49
& 25.22 & 15.91 & 30.77 & 19.41 & \textbf{40.00} & 25.24 & 37.69 & 23.78 \\
\textbf{\modelS}
& \best{46.01} & \best{35.54} & \best{55.00} & \best{52.30} & \best{70.38} & \best{67.17} & \best{29.23} & \best{17.82}
& \best{16.54} & \best{15.47} & \best{17.31} & \best{15.62} & 19.62 & \best{17.54} & \best{12.31} & \best{11.85}
& \best{30.43} & \best{29.94} & \best{31.54} & \best{29.92} & 38.08 & \best{36.15} & \best{38.46} & \best{36.32} \\
\addlinespace[0.05ex]
\midrule
\addlinespace[0.05ex]

VLM2VEC-2B
& 29.58 & 23.66 & 52.31 & 50.42 & 59.23 & 52.92 & 23.85 & 16.52
& 20.77 & 19.55 & \textbf{21.54} & \textbf{20.04} & 9.62 & 8.02 & \textbf{23.08} & \textbf{21.05}
& 20.87 & 19.87 & 26.15 & 25.04 & 29.23 & 28.50 & 21.15 & 20.31 \\
\textbf{\modelS}
& \best{47.42} & \best{39.51} & \best{61.15} & \best{58.78} & \best{71.15} & \best{66.14} & \best{26.92} & \best{18.35}
& \best{29.62} & \best{29.47} & 19.62 & 17.12 & \best{21.54} & \best{21.54} & 13.85 & 11.05
& \best{27.83} & \best{26.02} & \best{31.92} & \best{30.02} & \best{41.15} & \best{39.10} & \best{38.46} & \best{38.32} \\
\addlinespace[0.05ex]
\midrule
\addlinespace[0.05ex]

LLaVA1.6
& 32.39 & 25.78 & 58.06 & 55.45 & 63.08 & 52.46 & 28.46 & \textbf{20.09}
& 14.23 & 13.03 & 15.77 & 14.77 & \textbf{43.46} & \textbf{38.46} & \textbf{33.46} & \textbf{30.86}
& 22.61 & 20.85 & 25.00 & 23.89 & 37.31 & 35.69 & 22.69 & 21.05 \\
\textbf{\modelS}
& \best{46.01} & \best{37.54} & \best{58.30} & \best{55.50} & \best{70.77} & \best{63.01} & \best{28.85} & 20.02
& \best{28.08} & \best{27.98} & \best{23.08} & \best{22.00} & 33.46 & 30.46 & 24.62 & 22.62
& \best{36.52} & \best{35.05} & \best{28.85} & \best{27.88} & \best{41.54} & \best{40.84} & \best{39.23} & \best{38.52} \\
\addlinespace[0.05ex]
\midrule
\addlinespace[0.05ex]

Phi3-V
& 29.58 & 20.57 & 52.46 & 49.74 & 53.46 & 48.45 & 27.09 & 17.55
& \textbf{37.69} & \textbf{35.90} & 10.77 & 9.05 & 18.46 & 16.06 & 14.23 & 14.01
& 22.61 & 21.15 & \textbf{26.15} & \textbf{25.15} & \textbf{33.08} & \textbf{31.08} & 31.92 & 30.02 \\
\textbf{\modelS}
& \best{38.03} & \best{29.34} & \best{53.31} & \best{51.89} & \textbf{65.00} & \best{59.34} & \best{27.81} & \best{17.65}
& 29.23 & 28.01 & \best{26.54} & \best{25.95} & \best{29.23} & \best{27.52} & \best{20.77} & \best{18.77}
& \best{39.13} & \best{38.31} & 23.85 & 22.85 & 28.85 & 27.85 & \best{42.69} & \best{40.98} \\

\bottomrule
\end{tabular}}
\end{table*}

\section{Evaluation and Benchmarking}
In this section, we conduct experiments to evaluate \modelS, and analyze the empirical results.
\subsection{Experimental Setup}

\stitle{Task evaluation.}
All evaluations are conducted in a zero-shot setting. We report Hit@5 and NDCG@5\footnote{Hit@5 measures whether correct candidates are prioritized among the top results, while NDCG@5 evaluates the quality of the ranking within the top positions.} for all tasks. Note that our ranking candidates are intentionally constructed to be spatially or semantically plausible, reflecting real-world urban settings. 
%

\stitle{Baselines.}
We compare against strong vision–language models and multimodal embedding models to evaluate the effectiveness of our spatial data and two-stage training strategy. The VLM baselines include Qwen2-VL (2B/7B) \cite{wang2024qwen2}, Qwen2.5-VL (3B/7B) \cite{bai2025qwen2}, Phi-3-Vision-128K \cite{abdin2024phi3}, and LLaVA1.6-Mistral \cite{li2024llava}. Multimodal embedding models include GME-Qwen2VL-2B-Instruct \cite{zhang2024gme} and VLM2Vec-Qwen2VL-2B \cite{jiang2024vlm2vec}.

\stitle{Implementation and training details.}
All models are fine-tuned and evaluated using the \texttt{Swift} framework\footnote{\texttt{ms-swift} is a large-model fine-tuning and deployment framework provided by the ModelScope community. \url{https://github.com/modelscope/ms-swift}.}, with LoRA adapters and DeepSpeed for efficient finetuning. More details about finetuning can be found in Appendix~\ref{app:finetuning_uge}.

\vspace{-1mm}

\subsection{Benchmarking Performance on UGBench}
\stitle{Overall performance.}
As shown in Tables~\ref{tab:geo_retrieval} and \ref{tab:spatial_grounding_all}, spatial graph-enhanced fine-tuning consistently improves ranking performance, as measured by both Hit@5 and NDCG@5 across geolocation ranking, image retrieval, urban perception, and spatial grounding tasks. These gains are observed across all four cities, demonstrating strong generalization to held-out cities (Beijing and Paris), despite training only on New York and Singapore, with larger improvements generally seen in training cities. GME and VLM2Vec exhibit stronger zero-shot performance than their corresponding backbone VLM (i.e., Qwen2-VL-2B), reflecting the effectiveness of embedding-centric pretraining. However, the backbone Qwen2-VL benefits more substantially from \modelS\ than these general embedding models, indicating that explicit spatial grounding provides complementary gains beyond generic multimodal alignment. We also observe that Qwen-VL models benefit more substantially from \modelS\ than LLaVA1.6-Mistral and Phi-3-Vision. Notably, larger and newer Qwen models show proportionally larger improvements: \modelS\ fine-tuned on Qwen2.5-VL-7B achieves substantially greater gains than Qwen2-VL-7B, while Qwen2.5-VL-7B consistently outperforms Qwen2.5-VL-3B. This trend suggests that increased model capacity enables more effective utilization of structured spatial inductive bias. 

\begin{figure}[t]
    \centering
    \includegraphics[width=\columnwidth]{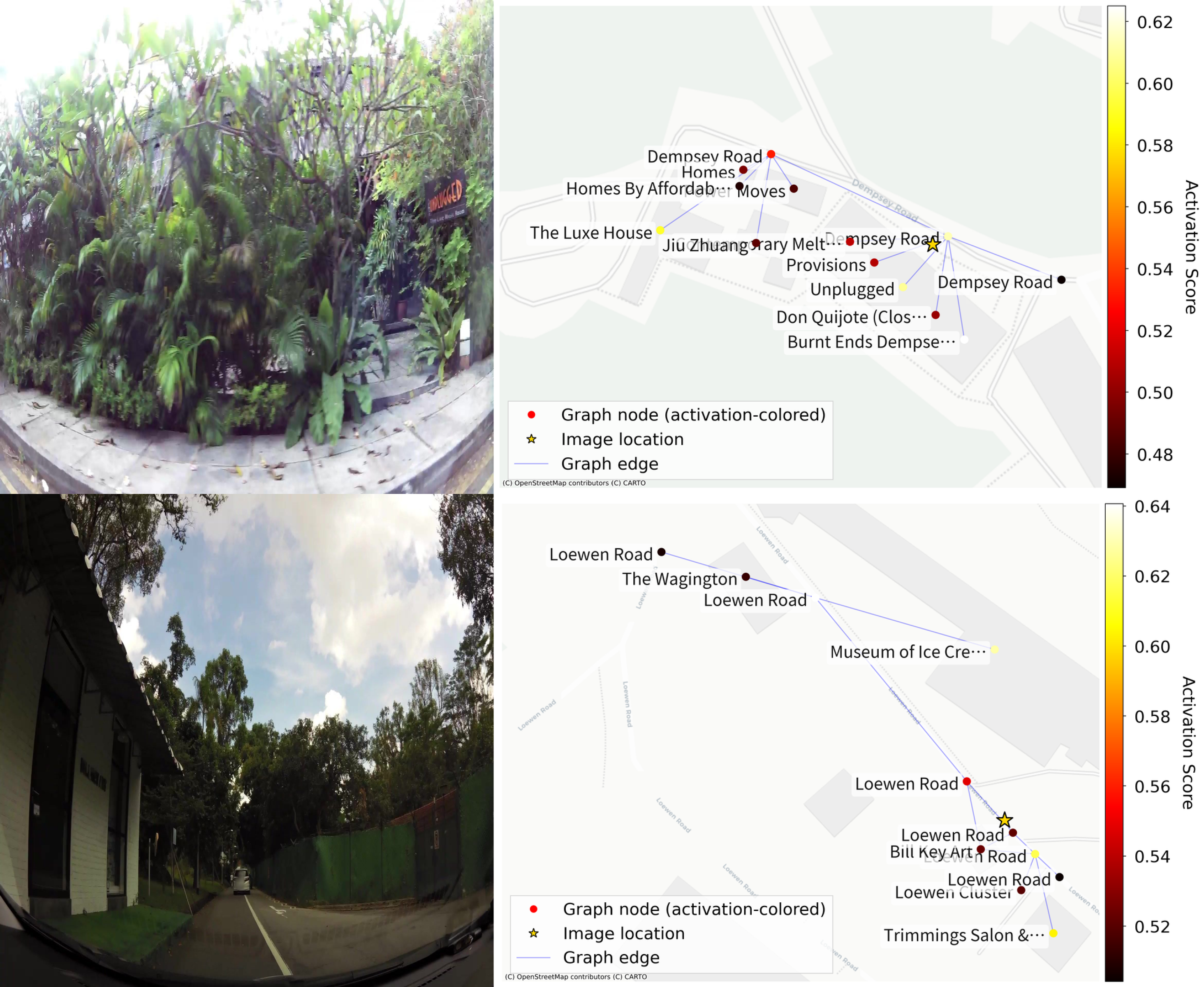}
    \vspace{-5mm}
    \caption{Spatial activation analysis for the depressing perception task. Two Singapore street views with similar greenery show different scores: the lower-depressing case (top) emphasizes activity-related POIs, while the higher-depressing case (bottom) focuses on nearby road segments.}
    \label{fig:depressing}
\end{figure}

\stitle{Analysis on individual tasks.}
Table~\ref{tab:geo_retrieval} shows that geolocation ranking exhibits consistent performance gains across all fine-tuned models (e.g. 30.38\% Hit@5 improvement on Singapore). This indicates that spatial knowledge learned from graphs is effectively internalized into the image embeddings, endowing them with strong spatial discriminative power for geographic localization. For urban perception tasks, \modelS\ consistently improves average performance across most VLM backbones especially on training cities, demonstrating that spatial grounding benefits high-level urban perception. Results for individual perception tasks are reported in Appendix~\ref{app:perception_tasks}. We further observe distinct spatial activation patterns for visually similar images. For example, Fig.~\ref{fig:depressing} contrasts two Singapore scenes with similar greenery but different depressiveness scores. In the lower-depressing case (top), \modelS\ built upon Qwen2.5-VL-7B attends to nearby commercial, residential, and recreational POIs, reflecting surrounding functional land uses and human activities that may mitigate perceived depressiveness, whereas in the higher-depressing case (bottom), attention shifts toward surrounding road segments. This contrast shows how spatial graphs complement visual cues by encoding urban functions and spatial relations beyond image content alone. For image retrieval, augmenting text queries with spatial graph information substantially improves retrieval and ranking performance over text-only queries (e.g., a 44\% Hit@5 gain on Singapore; Table~\ref{tab:geo_retrieval}), reflecting the benefit of aligning spatial graphs with textual and visual representations in UGE. 
For spatial grounding benchmark, Table~\ref{tab:spatial_grounding_all} shows that UGE consistently improves performance across cities except for Beijing on proximity-based tasks (e.g., nearest POI). This improvement largely stems from the structural encoding of proximity relationships in spatial graphs. In contrast, distance and distance-direction tasks show inconsistent performance. These queries involve fine-grained spatial concepts and explicit relational reasoning—such as directionality, relative orientation, and comparative distance—that may not be fully captured by the multimodal embeddings learned through contrastive training alone. This observation is consistent with prior findings in work on geographic question answering, which show that LLMs and VLMs often struggle with distance, direction, and other fine-grained spatial relations \cite{liao2024reasoning,mai2021geographic,yerramilli2025geochain}.

\subsection{Ablation Study}
\begin{figure}[t]
    \vspace{-0.8em}
    \centering
    \includegraphics[width=\columnwidth,height=0.28\textheight,keepaspectratio]{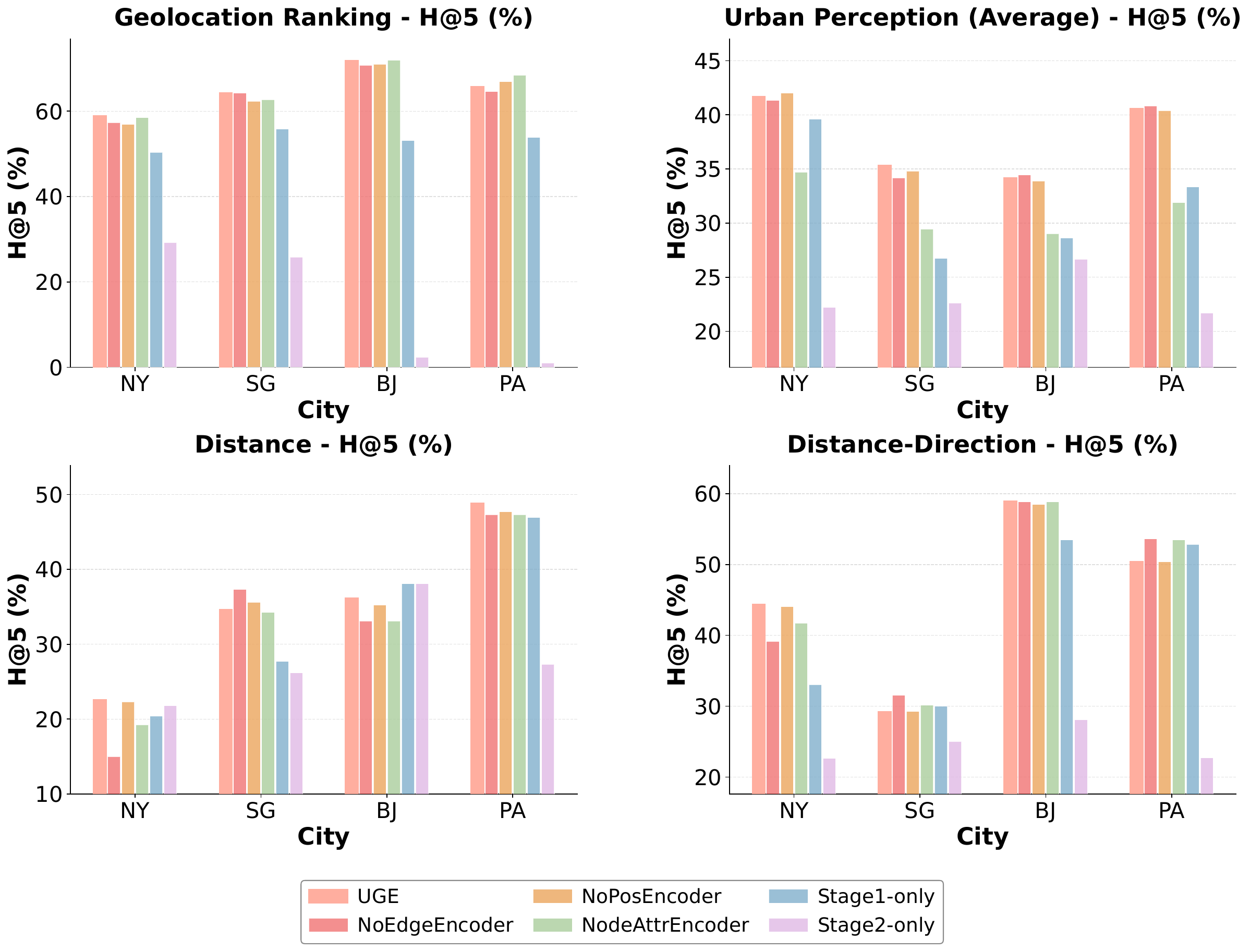}
    
    \vspace{-0.8em}
    \caption{Results of ablation study on \emph{Geolocation Ranking, Image Retrieval, Distance, Distance-Direction} tasks (Hit@5).}
    \vspace{-1em}
    \label{fig:ablation_four}
\end{figure}

We conduct an ablation study on \modelS\ with the Qwen2.5-VL-7B backbone to assess the impact of key design choices, by selectively modifying the spatial encoder or training curriculum. \NoEdgeEncoder\ removes edge features from the graph encoder. \NoPosEncoder\ removes positional encoding. \NodeAttrEncoder\ removes both edge features and positional encoding, reducing the graph to a node-only representation. \Stage\ denotes a model fine-tuned using only Stage~1 data, while \stage\ is trained directly on Stage~2 data without Stage~1 initialization. As shown in Figure~\ref{fig:ablation_four}, the full model (UGE) generally performs better across four tasks. Removing edge features or positional encoding leads to consistent but moderate degradation. The training curriculum has a larger effect: models trained with only Stage~1 or only Stage~2 data generally underperform the two-stage setup, highlighting the importance of progressively introducing spatial supervision. For metric spatial queries (i.e., distance and distance–direction), \NoEdgeEncoder\ occasionally performs better, suggesting that precisely modeling fine-grained quantitative and directional relations remains nontrivial within current graph-encoded representations. The complete results of the ablation study are included in the Appendix~\ref{app:ablation}.

\section{Conclusion, Outlook, and Broader Impact}
We presented \modelS\ framework for learning spatially grounded multimodal embeddings that aligned street-view images, text, and urban spatial graphs using the \dataset\ data pipeline. The proposed two-stage training strategy progressively introduced spatial reasoning cues and graph-conditioned learning to support stable and transferable representations. Evaluations on \benchmark\ showed that \modelS\ achieved substantial improvements on various urban tasks across most cities and model backbones. Notably, consistent gains in geolocation ranking and image retrieval suggested that spatial graphs could serve as effective priors for graph-guided geographic information retrieval across modalities. Moreover, our results highlight an important direction for future work: although graph-conditioned embeddings effectively capture relational and contextual spatial cues, improving the representation of fine-grained, metric, and directional spatial information remains essential for stronger spatial understanding and reasoning in multimodal models.

More broadly, our UGE framework offers a practical foundation for place-centric information retrieval, context-aware image and location search, geographical visual question answering, and AI travel assistants. Given that spatial graphs are easily accessible in real-world settings, we believe this work opens up meaningful directions for integrating explicit spatial context into multimodal models for reliable and interpretable place-centric AI applications.

\clearpage
\newpage

\section*{Acknowledgments}
We thank all collaborators for their valuable contributions, insightful discussions, and support throughout this work.

\bibliographystyle{ACM-Reference-Format}
\bibliography{references}

\clearpage
\newpage
\appendix
\section*{Appendices}
\renewcommand\thesubsection{\Alph{subsection}}
\renewcommand\thesubsubsection{\thesubsection.\arabic{subsection}}
\appendix
\UseRawInputEncoding
\subsection{Training Data Samples}
\label{app:data_samples}

\subsubsection{Stage 1: Instruction-Guided Image--Text Alignment}
\paragraph{Type 1: Image with Spatial Reasoning Path (SRP)}
This type injects spatial knowledge through language by pairing an image with a spatial reasoning path expressed in text.
\begin{lstlisting}
{
  "messages": [
    {
      "role": "user",
      "content": "<image> Describe if this viewpoint is reachable or the destination is reachable from the viewpoint."
    }
  ],
  "positive_messages": [
    [
      {
        "role": "user",
        "content": "... Spatial Reasoning Path:
        (539125490832143, nearest, Kallang Road) (2.0m, 342°(N)) ->
        (3.2m, 72°(E)) ->
        (Kallang Road, Intersection of Geylang Road and Kallang Road) ->
        (375.8m, 206°(SW)) ->
        (Geylang Road, near, Kallang Riverside Park North) ->
        (Kallang Riverside Park North, bounds, Kampong Bugis) ->
        (193.5m, 263°(W)) ->
        (Kampong Bugis, near, Banyan Tree & Ruin) (28.4m, 114°(SE)).
        Based on the spatial context, you can reach Banyan Tree & Ruin
        from the current image location at (103.8693, 1.3100)."
      }
    ]
  ],
  "label": 1,
  "images": [".../singapore/images/539125490832143.jpg"],
  "graphs": [],
  "pair_type": "stage1_image_only"
}
\end{lstlisting}

\paragraph{Type 2: Image-Level Captioning.}
This type aligns images with detailed semantic descriptions without explicit graph input.

\begin{lstlisting}[style=jsonsample]
{
  "messages": [
    {
      "role": "user",
      "content": "<image> Provide detailed descriptions of the image content."
    }
  ],
  "positive_messages": [
    [
      {
        "role": "user",
        "content": "The image captures a sunlit urban street scene on Broadway in Harlem, New York City, near the intersection of West 147th Street..."
      }
    ]
  ],
  "label": 1,
  "images": [".../newyork/images/2936878419934510.jpg"],
  "graphs": [],
  "pair_type": "stage1_image_only"
}
\end{lstlisting}

\subsubsection{Stage 2: Graph-Conditioned Spatial Embedding}

\paragraph{Type 1: Image--Graph with Spatial Reasoning Path.}
This type conditions learning on both the local spatial graph and an explicit spatial reasoning path, providing distance--direction cues and grounded place entities.

\begin{lstlisting}[style=jsonsample]
{
  "messages": [
    {
      "role": "user",
      "content": "<graph><image> Note on path notation: Triple indicates the decision point information in the path, the arrow '->' indicates the next step in the path, and parentheses (e.g., (50m, S)) immediately after a triple describe the distance and direction from the source to the target node. Based on the spatial context represented in graph and the spatial reasoning path, you can reach Ps 234 from the current location shown in the image with id 1360629940984808 at (-73.9196, 40.7724).\nHere are relevant place entities informations: {'1360629940984808': {'id': '1360629940984808', 'category': 'mapillary', 'coordinates': 'POINT(lon: -73.91961277, lat: 40.77240744)'}, '27 St': {'id': '538146', 'category': 'None', 'coordinates': 'LINE(...)'}, '20 Ave': {'id': '537435', 'category': 'None', 'coordinates': 'LINE(...)'}, 'Ps 234': {'id': '568292', 'category': 'None', 'coordinates': 'POINT(lon: -73.92307984, lat: 40.76702346)'}, ... }"
    },
    {
      "role": "assistant",
      "content": "You can reach Ps 234 from the current location shown in the image with id 1360629940984808 at (-73.9196, 40.7724)."
    }
  ],
  "images": ".../newyork/images/1360629940984808.jpg",
  "graphs": [
    ".../subgraphs/subgraph_1360629940984808.pkl"
  ],
  "summarization": "You can reach Ps 234 from the current location shown in the image with id 1360629940984808 at (-73.9196, 40.7724).",
  "image_coordinates": "(-73.9196, 40.7724)",
  "mapillary_node": "1360629940984808"
}
\end{lstlisting}

\paragraph{Type 2: Image--Graph Spatial Context Description.}
This type fuses visual content and the local spatial graph to produce a holistic description of the scene and its surrounding urban context.

\begin{lstlisting}[style=jsonsample]
{
  "messages": [
    {
      "role": "user",
      "content": "<graph><image> Use the image and graph together to describe the scene and its spatial context."
    }
  ],
  "positive_messages": [
    [
      {
        "role": "user",
        "content": "This image aligns closely with the spatial context of Bukit Batok, a planned residential town in  West District in Singapore. The location corresponds to a node near the complex crossing of Bukit Batok East Avenue 6 and Bukit Batok East Avenue 3..."
      }
    ]
  ],
  "label": 1,
  "images": [
    ".../singapore/images/1526294631371000.jpg"
  ],
  "graphs": [
    ".../subgraphs/mapillary_1526294631371000_subgraph.pkl"
  ],
  "pair_type": "stage2_fusion"
}
\end{lstlisting}

\subsection{Prompt for Spatial Context Caption Generation}
\label{app:scc}
To generate spatial context captions, we use a structured prompt that conditions image captioning on a localized urban spatial graph. The prompt consists of fixed instructions and a dynamic component, denoted as \texttt{\{subgraph\_desc\}}, which describes the spatial subgraph surrounding each street-view image.

\paragraph{Prompt Overview.}
The model is instructed to generate detailed captions for urban street images given a comprehensive spatial network context. The spatial context is provided as a textual description of a localized spatial graph derived from OpenStreetMap data.

\paragraph{Caption Generation Instructions.}
Given a single street-view image within the spatial network described above, the model is asked to produce:
\begin{itemize}[leftmargin=1.5em]
    \item \textbf{Detailed Image Caption}, describing image-specific visual features, including:
    locations, architectural characteristics, businesses and landmarks, vegetation and natural elements, visual indicators of acoustic environments (e.g., traffic or quiet settings), cues related to odors (e.g., vegetation, cafes, exhaust), and colors, materials, or textures influencing thermal and psychological perception.
    \item \textbf{Summarization of Image Features within Spatial Context}, focusing on visual elements that reflect or situate the image within its broader urban environment, such as neighborhood identity, street structure, proximity to other locations, characteristic urban design patterns, street signs or landmarks, and overall multi-sensory qualities affecting psychological wellbeing.
\end{itemize}

\paragraph{Additional Guidelines.}
The model is instructed to explicitly relate visual observations to entities and locations described in the spatial network, and to consider the interaction between visual, acoustic, and olfactory cues when characterizing the urban environment.

\paragraph{Spatial Context Caption Prompt.}
We use the following prompt to generate spatial context captions for each street-view image.
The prompt conditions caption generation on a localized urban spatial graph description.
The only dynamic component is \texttt{\{subgraph\_desc\}}, which is generated per image.

\begin{lstlisting}
You are an advanced vision model tasked with generating detailed captions
for urban street images within a comprehensive spatial network context.

# Comprehensive Spatial Context:

## Network Structure:
{subgraph_desc}

# Enhanced Caption Generation Instructions:

You will be provided with one street image from the spatial area described
in the Network Structure above. The image is situated within a spatial
context represented by graph information.

For the image, generate a detailed caption and then summarize the image
features based on the spatial context.

1. Detailed image caption:
   Describe the unique visual features of the image, including:
   - Locations
   - Distinctive architectural features
   - Notable businesses, signage, or landmarks
   - Vegetation, greenery, water features, or natural elements
   - Visual indicators of acoustic environments (traffic, construction,
     quiet settings)
   - Elements suggesting odors (vegetation, cafes, trash bins, traffic exhaust)
   - Colors, materials, and textures influencing thermal and psychological
     perception

2. Summarization of image features within spatial context:
   Focus on visual elements that reflect spatial context, including:
   - Clues indicating the neighborhood or area
   - Street features aligned with the network (crossings, width, type)
   - Visual indicators of proximity to other locations in the network
   - Architectural or urban design elements characteristic of the region
   - Street signs, direction indicators, or landmarks situating the image
   - Overall multi-sensory quality influencing psychological wellbeing

# Output Format:

Image:
<caption>

Summarization:
<summary>

# Instructions:
- Prefix captions with "Image:"
- End with a "Summarization:" section
- Connect visual observations to spatial connectivity patterns
- Refer to locations and entities in the network description
- Consider interactions of visual, acoustic, and olfactory cues
\end{lstlisting}

\subsection{Construction of Subgraph Description}
\label{app:sub_des}
To generate spatial context captions, we condition image captioning on a localized urban spatial graph. The prompt consists of fixed instructions and a dynamic component, denoted as \texttt{\{subgraph\_desc\}}, which encodes the spatial structure surrounding each street-view image.

\paragraph{Dynamic Subgraph Description (\texttt{\{subgraph\_desc\}}).}
The subgraph description is generated by the function

\texttt{\_create\_network\_description()} and is unique to each image. It reflects a localized OpenStreetMap-derived spatial graph centered at the image location. While the description follows a fixed structural template, its content (nodes, edges, attributes, distances, and directions) varies depending on the extracted subgraph.

\paragraph{Structure of the Subgraph Description.}
Each subgraph description is composed of four main sections:

\begin{itemize}[leftmargin=1.5em]
    \item \textbf{Network Overview.} A header summarizing the graph size, formatted as:
    \emph{``This network contains $N$ locations, with $C$ center nodes and $E$ connections.''}
    
    \item \textbf{Center Nodes.} One or more nodes corresponding to the image location(s) in the graph. Each center node is described by its identifier, semantic type, geometry, available attributes, geographic coordinates, and relative orientation.
    
    \item \textbf{Nearby Locations.} Up to ten nodes that are directly connected to the center node or lie within approximately 200 meters. These locations are sorted by distance and described using their name (if available), identifier, type, geometry, attributes, coordinates, and relative direction.
    
    \item \textbf{Locations Connected to Nearby Areas.} One-hop neighbors of the nearby locations, with up to twelve entries. These nodes follow the same descriptive format and provide extended neighborhood context.
    
    \item \textbf{Connections in the Network.} Explicit descriptions of edges, including:
    (i) connections between image locations and nearby areas, and
    (ii) connections among nearby areas.
    Each edge is annotated with relationship type (e.g., nearest, crossing, intersection), relative direction, distance, and crossing coordinates when applicable.
\end{itemize}

\paragraph{Node and Edge Attributes.}
Node descriptions may include attributes such as building use, historic district, architect, building type, address information (street, housenumber, postcode), planning area, district, city, and country. Edge descriptions include relationship type, bearing or direction, distance, and crossing identifiers or coordinates when available.

\paragraph{Example Subgraph Description.}
For illustration, a subgraph description may state that the image location lies near Orchard Road, ION Orchard, and Somerset MRT, and describe their relative distances, directions, and interconnections. In practice, the exact wording and attribute sets depend on the outputs of 

\texttt{\_extract\_node\_info()}, \texttt{\_get\_edge\_relationship\_description()}, and the available fields in the extracted node and edge GeoDataFrames.

\paragraph{Usage in Caption Generation.}
The generated \texttt{\{subgraph\_desc\}} is inserted into the spatial context prompt and provided alongside the image. This enables the model to generate captions and summaries that explicitly relate visual observations to surrounding streets, landmarks, connectivity patterns, and neighborhood structure, grounding image semantics within the broader urban spatial context.

\subsection{Spatial Reasoning Path Generation}
\label{app:srp}
For spatial grounding and navigation-style supervision, we generate
\emph{Spatial Reasoning Paths} (SRPs) that describe plausible routes from a
street-view image location to destination nodes in the surrounding urban
spatial graph. These paths are used as an additional dynamic component in
spatial reasoning prompts.

\paragraph{Path Generation Pipeline.}
Each spatial reasoning path is generated through a multi-stage pipeline
consisting of path discovery, filtering, formatting, and prompt integration.

\begin{itemize}[leftmargin=1.5em]
    \item \textbf{Path Discovery.} Starting from the image location (Mapillary
    node), we enumerate candidate paths to diverse destination nodes in the
    spatial graph using a breadth-first search over a precomputed adjacency
    structure.
    
    \item \textbf{Path Filtering and Selection.} Candidate paths are filtered
    based on hop distance, geometric distance thresholds, and node types (e.g.,
    excluding trivial intersections). To encourage diversity, paths of varying
    lengths are sampled across different hop distances.
    
    \item \textbf{Path Formatting.} Selected paths are converted into structured
    textual representations consisting of relation triples and
    distance--direction annotations, producing a compact and interpretable
    spatial reasoning path.
    
    \item \textbf{Training Data Integration.} The formatted spatial reasoning path is
    inserted into the user prompt as a dedicated \emph{Spatial Reasoning Path}
    section, together with a brief explanation of the path notation.
\end{itemize}

\paragraph{Path Discovery Strategy.}
Destination nodes are categorized by geometric type, including polygonal
entities (e.g., buildings or areas), linear entities (e.g., streets and roads),
and point entities (e.g., landmarks or intersections). A reverse breadth-first
search from the image location identifies reachable nodes within a maximum hop
limit. Candidates are prioritized by geometric distance from the image
location, with fallback to more distant nodes when necessary. For each selected
destination, the shortest valid path satisfying minimum hop constraints is
constructed.

\paragraph{Path Notation and Format.}
Each spatial reasoning path is expressed using a structured notation composed
of relation triples and distance--direction tuples. Triples encode decision
points along the route, while arrows indicate progression between successive
steps.

\begin{itemize}[leftmargin=1.5em]
    \item \textbf{Triples.} Each triple has the form
    $(\textit{source}, \textit{relation}, \textit{target})$, where source and
    target correspond to named OSM entities (e.g., streets, intersections, or
    landmarks). Relation types include street names, \texttt{nearest},
    \texttt{near}, \texttt{bounds}, \texttt{intersects},
    \texttt{intersection}, and 
    
    \texttt{complex\_crossing}.
    
    \item \textbf{Distance--Direction Tuples.} Each tuple specifies the distance
    (in meters) and bearing (in degrees with cardinal direction) between
    connected entities. Tuples may appear immediately after a triple or between
    steps to describe transitions along the path.
    
    \item \textbf{Arrow Separators.} The symbol \texttt{->} denotes progression
    to the next step in the route.
\end{itemize}

\paragraph{Illustrative Example.}
An example spatial reasoning path may begin at an intersection, move along a
street segment for several hundred meters in a given direction, pass through a
complex crossing, and terminate near a destination landmark. Such paths
explicitly expose relative distance, directionality, and connectivity, enabling
navigation-style spatial reasoning grounded in the underlying urban graph.

\paragraph{Role in Training and Evaluation.}
By translating graph-structured spatial relationships into compact linguistic
paths, SRPs provide a bridge between urban spatial graphs and language-based
supervision. They enable models to learn spatial concepts such as proximity,
direction, and multi-hop connectivity in a form compatible with
instruction-guided multimodal embedding learning.

\subsection{Benchmark Task Samples}
\label{app:benchmark_samples}

We provide representative examples from the four benchmark tasks used to evaluate spatially grounded urban embeddings.

\subsubsection{Geolocation Ranking}

This task evaluates fine-grained location discrimination by ranking candidate textual location descriptions given a street-view image.

\begin{lstlisting}
{
  "messages": [
    {
      "role": "user",
      "content": "<image> Where is this street view image taken?
                  Select the correct location:"
    }
  ],
  "images": [".../newyork/images/387138539252309.jpg"],
  "graphs": null,
  "ground_truth":
    "on Queens Blvd, near Intersection of 32 Pl and Queens Blvd,
     close to Sunnyside, in Sunnyside, Queens",
  "candidates": [
    "on Queens Blvd, near Intersection of 32 Pl and Queens Blvd,
     close to Sunnyside, in Sunnyside, Queens",
    "on Queens Blvd, near Intersection of 39 St and Queens Blvd,
     close to Sunnyside, in Sunnyside, Queens",
    "on Queens Plz S, near Intersection of Connector and Queens Plz,
     close to Long Island City, Queens",
    "... (remaining candidates omitted)"
  ]
}
\end{lstlisting}

\subsubsection{Image Retrieval with Spatial Graph Context}

This task evaluates text--graph--image alignment by retrieving the correct image given a spatial graph and textual spatial description.

\begin{lstlisting}
{
  "messages": [
    {
      "role": "user",
      "content": "<graph> With reference to the spatial graph,
                  pick the image captured near the intersection
                  of 3rd Avenue and East 10th Street in Manhattan."
    }
  ],
  "images": [],
  "graphs": [".../subgraphs/mapillary_497894054880543_subgraph.pkl"],
  "candidate_images": [
    ".../images/526798132036513.jpg",
    ".../images/2609168706058382.jpg",
    ".../images/497894054880543.jpg",
    "... (remaining candidates omitted)"
  ],
  "ground_truth_idx": 12,
  "ground_truth":
    ".../images/497894054880543.jpg"
}
\end{lstlisting}

\subsubsection{Urban Perception Prediction (Depressiveness)}

This task evaluates high-level urban perception by ranking discretized perception labels derived from human annotations.

\begin{lstlisting}
{
  "messages": [
    {
      "role": "user",
      "content": "<image> What is the perception of depressing
                  for this urban location?"
    }
  ],
  "images": [".../beijing_paris_tokyo/285075470021284.jpg"],
  "graphs": [],
  "ground_truth":
    "Less depressing, Depressing Score: 4.5",
  "candidates": [
    "Not depressing, Depressing Score: 2.5",
    "Less depressing, Depressing Score: 4.5",
    "Moderately depressing, Depressing Score: 6.0",
    "Very depressing, Depressing Score: 9.0",
    "... (remaining candidates omitted)"
  ]
}
\end{lstlisting}

\subsubsection{Spatial Grounding: Distance Query}

This task evaluates metric spatial reasoning by predicting distances between entities using both image and graph context.

\begin{lstlisting}
{
  "messages": [
    {
      "role": "user",
      "content": "<image><graph>
                  How far is the nearest finance
                  from the closest amenity?"
    }
  ],
  "images": [".../paris/images/4321631924514461.jpg"],
  "graphs": [".../paris/subgraphs/subgraph_4321631924514461.pkl"],
  "ground_truth": "33 meters",
  "candidates": [
    "33 meters",
    "101 meters",
    "304 meters",
    "506 meters",
    "800 meters",
    "1206 meters"
  ]
}
\end{lstlisting}

\subsection{Details of Finetuning UGE}
\label{app:finetuning_uge}
Stage~1 training uses batch sizes in \{8, 9, 10, 12\} for image--text contrastive learning, while Stage~2 adopts smaller batch sizes \{3, 4\} to accommodate graph-conditioned inputs. The graph encoder uses two GNN layers with edge-aware attention, 64-dimensional edge encoding and spatial positional encoding \cite{lee2024latent}, producing node embeddings projected to the VLM hidden dimension. To ensure stable training when introducing the graph modality, we apply differentiated learning rates, updating the graph encoder at $5 \times 10^{-5}$ while scaling VLM parameter updates to 10\% of this rate. We use different hardware configurations across training stages and model families. Stage~2 fine-tuning of the LLaVA1.6-Mistral uses 4 NVIDIA A800 GPUs for 3 epochs ($\approx$1 day and 12 hours), while all other models are trained on 4 NVIDIA RTX 5000 GPUs for 5 epochs (12--18 hours).

\subsection{Results of Each Task in Urban Perception Benchmark}
\label{app:perception_tasks}
Table~\ref{tab:perception1} and Table~\ref{tab:perception2} presents per-task results for the urban perception benchmark which show inconsistent performance across different attributes of urban perceptions.
\begin{table*}[!t]
\centering
\caption{Urban perception results on \textit{Safe}, \textit{Wealthy}, and \textit{Lively} (H@5 / N@5, \%).}
\label{tab:perception1}
\footnotesize
\setlength{\tabcolsep}{3.5pt}
\renewcommand{\arraystretch}{1.05}
\resizebox{0.98\textwidth}{!}{%
\begin{tabular}{l cc|cc|cc|cc|cc|cc|cc|cc|cc|cc|cc|cc}
\toprule
\textbf{Model} & \multicolumn{8}{c}{\textbf{Safe}} & \multicolumn{8}{c}{\textbf{Wealthy}} & \multicolumn{8}{c}{\textbf{Lively}} \\
\cmidrule(lr){2-9} \cmidrule(lr){10-17} \cmidrule(lr){18-25}
& \multicolumn{2}{c}{NY} & \multicolumn{2}{c}{SG} & \multicolumn{2}{c}{BJ} & \multicolumn{2}{c}{PA}
& \multicolumn{2}{c}{NY} & \multicolumn{2}{c}{SG} & \multicolumn{2}{c}{BJ} & \multicolumn{2}{c}{PA}
& \multicolumn{2}{c}{NY} & \multicolumn{2}{c}{SG} & \multicolumn{2}{c}{BJ} & \multicolumn{2}{c}{PA} \\
\cmidrule(lr){2-3} \cmidrule(lr){4-5} \cmidrule(lr){6-7} \cmidrule(lr){8-9}
\cmidrule(lr){10-11} \cmidrule(lr){12-13} \cmidrule(lr){14-15} \cmidrule(lr){16-17}
\cmidrule(lr){18-19} \cmidrule(lr){20-21} \cmidrule(lr){22-23} \cmidrule(lr){24-25}
& Hit & NDCG & Hit & NDCG & Hit & NDCG & Hit & NDCG
& Hit & NDCG & Hit & NDCG & Hit & NDCG & Hit & NDCG
& Hit & NDCG & Hit & NDCG & Hit & NDCG & Hit & NDCG \\
\midrule

Qwen2-VL-2B
& 32.90 & 20.00 & 35.99 & 21.93 & 45.24 & 26.44 & 42.93 & 24.91
& 30.59 & 16.41 & 21.34 & 11.72 & \textbf{26.99} & \textbf{15.63} & \textbf{31.11} & \textbf{20.46}
& 11.05 & 6.66 & 17.99 & 10.94 & \textbf{15.42} & \textbf{8.51} & 5.66 & 3.44 \\
\textbf{\modelS}
& \best{33.16} & \best{20.40} & \best{36.76} & \best{22.98} & \best{48.84} & \best{28.33} & \best{43.44} & \best{25.61}
& \best{32.90} & \best{17.83} & \best{31.88} & \best{19.33} & {24.42} & {12.83} & {20.11} & {15.75} & \best{31.11} & \best{18.75} & \best{18.61} & \best{11.42} & 12.80 & 6.95 & \best{31.62} & \best{17.84} \\
\addlinespace[0.05ex]
\midrule
\addlinespace[0.05ex]
Qwen2-VL-7B
& 19.02 & 10.74 & 17.22 & 11.08 & 36.25 & 24.20 & 25.71 & 13.31
& 13.62 & 6.27 & 24.42 & 12.90 & 27.25 & 16.62 & 20.05 & 10.34
& 27.51 & 14.93 & 19.79 & 9.05 & \textbf{19.56} & \textbf{15.85} & \textbf{33.68} & \textbf{18.44} \\
\textbf{\modelS}
& \best{24.68} & \best{12.75} & \best{26.99} & \best{15.41} & \best{40.62} & \best{21.39} & \best{29.05} & \best{15.99}
& \best{34.96} & \best{20.08} & \best{30.33} & \best{17.78} & \best{32.39} & \best{19.11} & \best{26.22} & \best{15.47}
& \best{31.11} & \best{16.61} & \best{31.62} & \best{18.29} & 17.99 & 10.69 & 20.05 & 12.07 \\
\addlinespace[0.05ex]
\midrule
\addlinespace[0.05ex]
Qwen2.5-VL-3B
& 27.06 & 14.59 & 18.26 & 10.72 & 7.12 & 3.67 & 8.23 & 4.11
& 11.23 & 5.39 & 15.94 & 8.58 & 15.42 & 9.06 & 26.48 & 14.93
& \textbf{39.07} & \textbf{26.70} & \textbf{34.70} & \textbf{21.89} & 45.24 & 20.49 & \textbf{45.24} & \textbf{29.13} \\
\textbf{\modelS}
& \best{27.51} & \best{15.30} & \best{25.45} & \best{17.02} & \best{16.20} & \best{11.00} & \best{20.05} & \best{13.22}
& \best{32.13} & \best{18.12} & \best{43.19} & \best{25.68} & \best{33.93} & \best{20.56} & \best{39.07} & \best{21.28}
& 20.33 & 12.08 & 30.33 & 17.90 & \best{32.90} & \best{22.55} & 37.02 & 21.25 \\
\addlinespace[0.05ex]
\midrule
\addlinespace[0.05ex]
Qwen2.5-VL-7B
& 27.25 & 16.19 & 28.28 & 17.18 & 35.48 & 22.85 & 35.73 & 20.61
& 27.51 & 16.25 & 36.25 & 21.27 & 35.22 & 20.17 & 27.25 & 14.88
& 18.25 & 9.61 & 30.08 & 17.88 & 24.94 & 15.81 & 25.19 & 13.73 \\
\textbf{\modelS}
& \best{37.02} & \best{25.30} & \best{31.11} & \best{20.83} & \best{36.76} & \best{24.09} & \best{31.36} & \best{20.03}
& \best{35.48} & \best{22.26} & \best{40.87} & \best{28.87} & \best{35.28} & \best{23.80} & \best{45.50} & \best{31.11}
& \best{60.93} & \best{42.40} & \best{39.85} & \best{26.77} & \best{37.08} & \best{24.23} & \best{48.59} & \best{33.56} \\
\addlinespace[0.05ex]
\midrule
\addlinespace[0.05ex]
GME-2B
& 26.22 & 12.56 & 29.82 & 15.06 & 37.02 & 17.60 & 27.51 & 13.17
& \textbf{25.45} & \textbf{12.17} & 29.31 & 14.86 & 21.59 & 10.55 & 14.14 & 6.73
& \textbf{23.19} & \textbf{14.33} & 23.39 & 11.67 & \textbf{19.28} & \textbf{9.58} & \textbf{24.68} & \textbf{11.97} \\
\textbf{\modelS}
& \best{33.23} & \best{17.41} & \best{39.07} & \best{23.75} & \best{22.62} & \best{13.48} & \best{24.16} & \best{13.23}
& 22.37 & 12.60 & \best{33.68} & \best{21.13} & \best{26.99} & \best{17.50} & \best{21.59} & \best{12.33}
& \best{9.51} & \best{4.35} & \best{7.97} & \best{4.69} & 13.11 & 6.83 & 11.83 & 6.72 \\
\addlinespace[0.05ex]
\midrule
\addlinespace[0.05ex]
VLM2Vec-2B
& 8.48 & 4.23 & 11.83 & 5.98 & 14.91 & 6.81 & 16.71 & 8.29
& 13.88 & 6.61 & 5.40 & 2.37 & 12.60 & 5.75 & 5.66 & 2.55
& 12.60 & 5.63 & 14.14 & 6.31 & 11.21 & 4.86 & 9.25 & 4.24 \\
\textbf{\modelS}
& \best{41.39} & \best{23.55} & \best{17.99} & \best{10.81} & \best{38.82} & \best{22.66} & \best{29.05} & \best{16.62}
& \best{22.11} & \best{12.94} & \best{16.71} & \best{9.13} & \best{20.57} & \best{10.84} & \best{15.17} & \best{8.78}
& \best{37.28} & \best{21.74} & \best{25.71} & \best{14.93} & \best{11.31} & \best{7.46} & \best{19.03} & \best{10.41} \\
\addlinespace[0.05ex]
\midrule
\addlinespace[0.05ex]
LLaVA-1.7
& 35.73 & 23.84 & \textbf{27.25} & \textbf{16.05} & \textbf{28.79} & \textbf{17.92} & \textbf{25.45} & \textbf{14.64}
& 29.07 & 16.89 & \textbf{36.50} & 2\textbf{0.72} & \textbf{32.39} & \textbf{18.88} & \textbf{21.08} & \textbf{11.92}
& 56.82 & 33.77 & 33.68 & \textbf{21.89} & \textbf{37.53} & 21.81 & \textbf{53.98} & \textbf{28.89} \\
\textbf{\modelS}
& \best{35.90} & \best{23.98} & 21.85 & 11.81 & 19.79 & 11.23 & 20.31 & 11.75
& \best{32.65} & \best{18.90} & 27.51 & 17.19 & 24.16 & 14.29 & 17.48 & 9.89
& \best{62.99} & \best{39.25} & \best{34.70} & 20.93 & 36.50 & \best{22.59} & 40.87 & 24.65 \\
\addlinespace[0.05ex]
\midrule
\addlinespace[0.05ex]
Phi3-V
& 31.36 & 17.63 & 30.34 & 17.18 & 26.48 & 13.79 & 24.68 & 13.90
& \textbf{31.11} & \textbf{18.35} & \textbf{41.65} & \textbf{24.94} & 28.54 & 17.59 & 28.02 & 15.98
& \textbf{38.05} & \textbf{19.72} & \textbf{39.07} & \textbf{23.64} & \textbf{37.79} & \textbf{21.83} & \textbf{41.90} & \textbf{23.30} \\
\textbf{\modelS}
& \best{36.51} & \best{20.25} & \best{30.85} & \best{18.53} & \best{43.70} & \best{25.85} & \best{35.99} & \best{21.80}
& 26.99 & 15.26 & 30.59 & 17.83 & \best{29.05} & \best{18.18} & \best{30.85} & \best{18.23}
& 21.85 & 10.55 & 22.11 & 12.17 & 22.37 & 13.45 & 23.91 & 14.38 \\
\bottomrule
\end{tabular}}
\end{table*}
\begin{table*}[!t]
\centering
\caption{Urban perception results on \textit{Depressing}, \textit{Beautiful}, and \textit{Boring} (H@5 / N@5, \%).}
\label{tab:perception2}
\footnotesize
\setlength{\tabcolsep}{3.5pt}
\renewcommand{\arraystretch}{1.05}
\resizebox{0.98\textwidth}{!}{
\begin{tabular}{l cc|cc|cc|cc|cc|cc|cc|cc|cc|cc|cc|cc}
\toprule
\textbf{Model} & \multicolumn{8}{c}{\textbf{Depressing}} & \multicolumn{8}{c}{\textbf{Beautiful}} & \multicolumn{8}{c}{\textbf{Boring}} \\
 & \multicolumn{2}{c}{NY} & \multicolumn{2}{c}{SG} & \multicolumn{2}{c}{BJ} & \multicolumn{2}{c}{PA}
 & \multicolumn{2}{c}{NY} & \multicolumn{2}{c}{SG} & \multicolumn{2}{c}{BJ} & \multicolumn{2}{c}{PA}
 & \multicolumn{2}{c}{NY} & \multicolumn{2}{c}{SG} & \multicolumn{2}{c}{BJ} & \multicolumn{2}{c}{PA} \\
 & Hit & NDCG & Hit & NDCG & Hit & NDCG & Hit & NDCG
 & Hit & NDCG & Hit & NDCG & Hit & NDCG & Hit & NDCG
 & Hit & NDCG & Hit & NDCG & Hit & NDCG & Hit & NDCG \\
\midrule
Qwen2-VL-2B & 7.46 & 3.30 & 12.34 & 6.60 & \textbf{15.98} & \textbf{8.09} & 15.68 & 8.09 & \textbf{41.39} & \textbf{23.47} & 27.74 & 17.18 & 21.85 & 11.40 & \textbf{35.73} & \textbf{23.80} & 13.11 & 8.72 & 11.83 & 6.66 & \textbf{15.71} & \textbf{9.76} & 13.37 & \textbf{8.38} \\
\textbf{\modelS} & \best{36.25} & \best{21.08} & \best{14.40} & \best{8.56} & 11.57 & 6.58 & \best{20.82} & \best{12.34} & 36.50 & 21.28 & \best{34.29} & \best{21.65} & \best{31.88} & \best{18.11} & 20.06 & 11.50 & \best{18.87} & \best{12.26} & \best{25.96} & \best{14.00} & 13.88 & 8.37 & \best{22.37} & \best{13.46} \\
\addlinespace[0.05ex]
\midrule
\addlinespace[0.05ex]
Qwen2-VL-7B & 12.34 & 5.40 & 28.53 & 14.68 & \textbf{44.99} & 
\textbf{29.10} & 18.51 & 7.79 & 16.20 & 7.70 & \textbf{29.05} & \textbf{19.05} & 10.03 & 5.10 & 22.37 & 12.51 & 12.08 & 5.64 & \textbf{26.99} & 14.83 & \textbf{26.22} & \textbf{14.76} & 14.40 & 6.59 \\
\textbf{\modelS} & \best{43.96} & \best{26.61} & \best{37.79} & \best{22.93} & 19.28 & 10.52 & \best{32.13} & \best{19.50} & \best{25.96} & \best{15.25} & 24.42 & 14.58 & \best{27.76} & \best{16.25} & \best{25.45} & \best{15.63} & \best{31.11} & \best{16.82} & 25.96 & \best{15.83} & 16.45 & 9.09 & \best{34.45} & \best{18.68} \\
\addlinespace[0.05ex]
\midrule
\addlinespace[0.05ex]
Qwen2.5-VL-3B & 14.50 & \best{12.27} & 11.57 & \textbf{6.26} & \textbf{14.65} & \textbf{7.94} & 9.00 & 4.77 & 30.59 & 17.94 & \textbf{31.36} & \textbf{18.48} & 11.57 & 8.21 & 28.02 & 16.55 & 24.42 & 12.83 & 11.57 & 7.26 & 6.94 & 3.51 & 6.68 & 4.06 \\
\textbf{\modelS} & \best{15.17} & 11.09 & \best{12.85} & 5.85 & 11.05 & 5.38 & \best{12.85} & \best{6.74} & \best{32.39} & \best{18.33} & 26.99 & 15.60 & \best{24.16} & \best{15.22} & \best{35.48} & \best{20.57} & \best{25.36} & \best{13.58} & \best{18.25} & \best{11.47} & \best{10.28} & \best{7.04} & \best{14.40} & \best{9.60} \\
\addlinespace[0.05ex]
\midrule
\addlinespace[0.05ex]
Qwen2.5-VL-7B & 17.48 & 7.18 & 25.96 & 13.50 & 35.99 & 22.54 & 19.79 & 10.33 & 27.51 & 15.34 & 29.31 & 15.60 & 23.39 & 15.07 & 28.79 & 17.02 & 4.63 & 2.51 & 7.20 & 4.56 & 7.46 & 4.08 & 6.17 & 3.35 \\
\textbf{\modelS} & \best{50.13} & \best{33.53} & \best{46.02} & \best{31.95} & \best{47.56} & \best{33.11} & \best{50.90} & \best{36.22} & \best{35.22} & \best{21.25} & \best{32.13} & \best{19.81} & \best{28.79} & \best{18.60} & \best{28.89} & \best{17.12} & \best{31.36} & \best{17.98} & \best{22.11} & \best{13.86} & \best{19.54} & \best{12.42} & \best{38.30} & \best{22.81} \\
\addlinespace[0.05ex]
\midrule
\addlinespace[0.05ex]
GME-2B & 33.16 & 16.83 & 34.45 & 16.59 & \textbf{30.85} & \textbf{15.08} & \textbf{33.16} & \textbf{16.43} & \textbf{25.45} & 12.44 & 21.08 & 9.86 & 22.11 & 10.40 & 21.08 & 10.45 & \textbf{34.70} & \textbf{17.23} & \textbf{31.36} & \textbf{15.07} & \textbf{49.36} & \textbf{24.58} & \textbf{38.82} & \textbf{18.87} \\
\textbf{\modelS} & \best{33.23} & \best{17.41} & \best{39.07} & \best{23.75} & 22.62 & 13.48 & 24.16 & 13.23 & 22.37 & \best{12.60} & \best{33.68} & \best{21.13} & \best{26.99} & \best{17.50} & \best{21.59} & \best{12.33} & 29.51 & 14.35 & 27.97 & 10.69 & 13.11 & 6.83 & 11.83 & 6.72 \\
\addlinespace[0.05ex]
\midrule
\addlinespace[0.05ex]
VLM2VEC-2B & 21.85 & 10.10 & 23.65 & 10.83 & 19.79 & 9.17 & 21.34 & 10.09 & 20.05 & 10.14 & 32.90 & 16.74 & 19.28 & 9.25 & 21.59 & 10.65 & \textbf{32.73} & \textbf{17.34} & 23.65 & 10.30 & \textbf{35.73} & \textbf{17.34} & \textbf{39.59} & \textbf{19.43} \\
\textbf{\modelS} & \best{28.28} & \best{14.10} & \best{29.31} & \best{16.71} & \best{20.31} & \best{11.84} & \best{26.74} & \best{15.38} & \best{33.93} & \best{20.99} & \best{34.96} & \best{22.24} & \best{29.82} & \best{18.55} & \best{31.36} & \best{18.14} & 23.14 & 12.04 & \best{26.74} & \best{15.48} & 25.74 & 10.63 & 29.05 & 15.92 \\
\addlinespace[0.05ex]
\midrule
\addlinespace[0.05ex]
Llava 1.7 & 15.68 & 7.07 & 15.68 & 8.84 & 17.22 & 10.81 & 16.71 & 9.64 & \textbf{31.11} & \textbf{19.45} & 19.54 & 10.78 & \textbf{25.45} & \textbf{15.46} & 23.65 & 13.59 & 5.66 & 3.98 & 7.71 & 4.84 & 6.43 & 3.64 & 6.43 & 3.79 \\
\textbf{\modelS} & \best{25.68} & \best{12.73} & \best{17.79} & \best{9.53} & \best{17.74} & \best{10.32} & \best{17.99} & \best{10.02} & 24.42 & 14.24 & \best{28.53} & \best{16.96} & 15.94 & 9.34 & \best{24.88} & \best{14.66} & \best{31.62} & \best{18.06} & \best{16.71} & \best{8.93} & \best{25.96} & \best{15.67} & \best{23.14} & \best{11.53} \\
\addlinespace[0.05ex]
\midrule
\addlinespace[0.05ex]
Phi3-V & 6.17 & 2.64 & 18.25 & 8.69 & 14.65 & 6.95 & 10.28 & 5.54 & \textbf{38.82} & \textbf{23.09} & \textbf{37.02} & \textbf{21.91} & 23.39 & 15.02 & \textbf{34.45} & \textbf{20.28} & 17.48 & 8.80 & \textbf{24.42} & \textbf{12.93} & 13.88 & 7.25 & 16.45 & 9.37 \\
\textbf{\modelS} & \best{38.56} & \best{24.41} & \best{23.65} & \best{13.74} & \best{22.62} & \best{14.42} & \best{29.05} & \best{17.85} & 29.56 & 16.83 & 14.91 & 8.67 & \best{26.74} & \best{16.65} & 26.99 & 14.30 & \best{17.85} & \best{9.72} & 17.74 & 9.84 & \best{21.59} & \best{11.38} & \best{19.28} & \best{9.89} \\
\bottomrule
\end{tabular}}
\end{table*}

\subsection{Results of \emph{Nearest Street} Task in Spatial Grounding Benchmark}
\label{app:nearest_street}

Table~\ref{tab:spatial_grounding_nearest_street_col} presents per-city results for the \emph{Nearest Street} task in the spatial grounding benchmark. Generally, UGE models perform better than the baselines, especially in training cities.

\noindent
\begin{minipage}{\columnwidth}
\captionsetup{type=table}
\captionof{table}{Results of \emph{Nearest Street} task (H@5 / N@5, \%).}
\vspace{-0.3em}
\label{tab:spatial_grounding_nearest_street_col}
\footnotesize
\setlength{\tabcolsep}{1.8pt}
\renewcommand{\arraystretch}{0.95}
\centering
\resizebox{\columnwidth}{!}{%
\begin{tabular}{l cc cc cc cc}
\toprule
\textbf{Model} & \multicolumn{2}{c}{NY} & \multicolumn{2}{c}{SG} & \multicolumn{2}{c}{BJ} & \multicolumn{2}{c}{PA} \\
\cmidrule(lr){2-3}\cmidrule(lr){4-5}\cmidrule(lr){6-7}\cmidrule(lr){8-9}
& H@5 & N@5 & H@5 & N@5 & H@5 & N@5 & H@5 & N@5 \\
\midrule
Qwen2-VL-2B   & 40.29 & 25.15 & 27.83 & 15.84 & 18.17 & 10.98 & 20.00 & 11.78 \\
\textbf{\modelS} & \best{42.09} & \best{28.68} & \best{49.83} & \best{31.63} & \best{32.83} & \best{20.45} & \best{39.67} & \best{26.76} \\
\midrule
Qwen2-VL-7B   & 34.17 & 19.62 & 25.33 & 14.35 & 10.33 & 7.55  & 56.33 & 34.91 \\
\textbf{\modelS} & \best{44.60} & \best{32.43} & \best{55.00} & \best{37.19} & \best{51.00} & \best{30.41} & \best{49.00} & \best{32.94} \\
\midrule
Qwen2.5-VL-3B & 25.54 & 14.71 & 23.00 & 13.67 & 39.33 & 29.37 & 34.17 & 21.67 \\
\textbf{\modelS} & \best{37.41} & \best{23.40} & \best{33.67} & \best{19.57} & \best{50.67} & \best{33.87} & \best{49.83} & \best{31.93} \\
\midrule
Qwen2.5-VL-7B & 37.41 & 22.59 & 30.50 & 15.88 & 25.00 & 19.02 & 20.00 & 12.19 \\
\textbf{\modelS} & \best{38.13} & \best{26.18} & \best{55.17} & \best{37.71} & \best{41.90} & \best{32.04} & \best{52.20} & \best{34.30} \\
\midrule
GME-2B        & 36.69 & 18.74 & 29.83 & 14.55 & \textbf{48.67} & \textbf{23.90} & 27.33 & 13.45 \\
\textbf{\modelS} & \best{40.29} & \best{24.78} & \best{45.67} & \best{29.29} & 22.33 & 11.25 & \best{48.17} & \best{28.77} \\
\midrule
VLM2VEC-2B    & 24.10 & 13.03 & 22.50 & 12.50 & 23.17 & 15.30 & 8.17  & 4.20  \\
\textbf{\modelS} & \best{40.65} & \best{27.40} & \best{53.83} & \best{35.41} & \best{28.00} & \best{17.05} & \best{34.83} & \best{23.38} \\
\midrule
LLaVA1.6      & 22.66 & 12.98 & 21.83 & 12.35 & \textbf{53.83} & \textbf{31.96} & 30.17 & 16.21 \\
\textbf{\modelS} & \best{47.84} & \best{34.91} & \best{69.33} & \best{46.82} & 33.83 & 20.56 & \best{45.00} & \best{31.29} \\
\midrule
Phi3-V        & 25.54 & 15.10 & 25.17 & 14.04 & \textbf{36.33} & \textbf{20.77} & 34.00 & 20.29 \\
\textbf{\modelS} & \best{40.29} & \best{28.03} & \best{48.17} & \best{31.99} & 32.50 & 17.71 & \best{45.50} & \best{29.18} \\
\bottomrule
\end{tabular}}
\end{minipage}


\subsection{Additional Ablation Studies}
\label{app:ablation}

This section presents ablation results that disentangle the contributions of individual components in our two-stage training pipeline across geolocation ranking, image retrieval, urban perception, and spatial grounding tasks.
For geolocation ranking, Figure \ref{fig:ablation_geo_rank} shows that the full model (UGE) generally outperforms its variants, indicating that both the complete spatial encoder and the two-stage training curriculum contribute to more discriminative embeddings for location detection. Figure \ref{fig:ablation_image_retrieval} reveals that although UGE generally performs better in terms of Hit@5, fine-tuning using only Stage~1 data outperforms other model variants in ranking quality, especially in held-out cities. 
Figure \ref{fig:ablation_urban_perception} evaluates high-level urban perception, reporting average performance across perceptual attributes and shows that UGE yeild the best overal performance.
Figures \ref{fig:ablation_nearest_street} and \ref{fig:ablation_nearest_poi} show that variants removing edge features or positional encodings (\NoEdgeEncoder, \NoPosEncoder, \NodeAttrEncoder) exhibit moderate yet systematic performance degradation. In contrast, models trained without the two-stage curriculum (\Stage, \stage) display more inconsistent gains across different city contexts. This suggests that the alignment introduced in Stage~1 already enables the VLM embeddings to capture certain proximity-related spatial cues, even without explicit graph-conditioned supervision.
Figures \ref{fig:ablation_distance} and \ref{fig:ablation_distance_direction} present ablation results on distance-aware and compositional spatial queries. Across model variants, performance gains are less consistent than in proximity-based tasks. Variants without edge encoding occasionally outperform the full model on these metric queries, reflecting the challenge of encoding precise quantitative and directional relations within the current graph-based embedding framework. These results further highlight the limits of existing spatial representations for metric reasoning, despite their effectiveness for proximity-based grounding.

\begin{center}
\captionsetup{type=figure}
\includegraphics[width=\columnwidth,height=0.28\textheight,keepaspectratio]{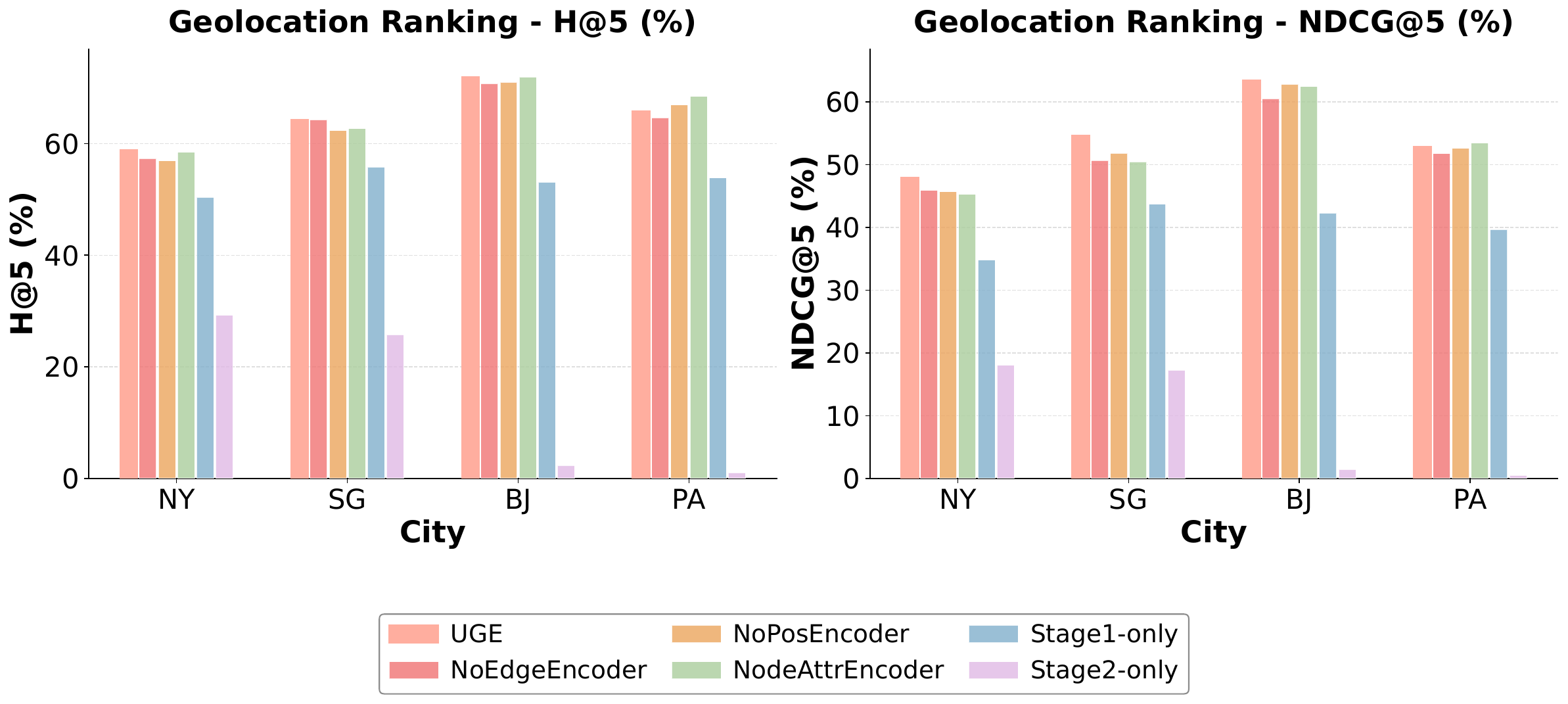}
\captionof{figure}{Results of ablation study on \emph{Geolocation Ranking}.}
\label{fig:ablation_geo_rank}
\end{center}

\begin{center}
\captionsetup{type=figure}
\includegraphics[width=\columnwidth,height=0.28\textheight,keepaspectratio]{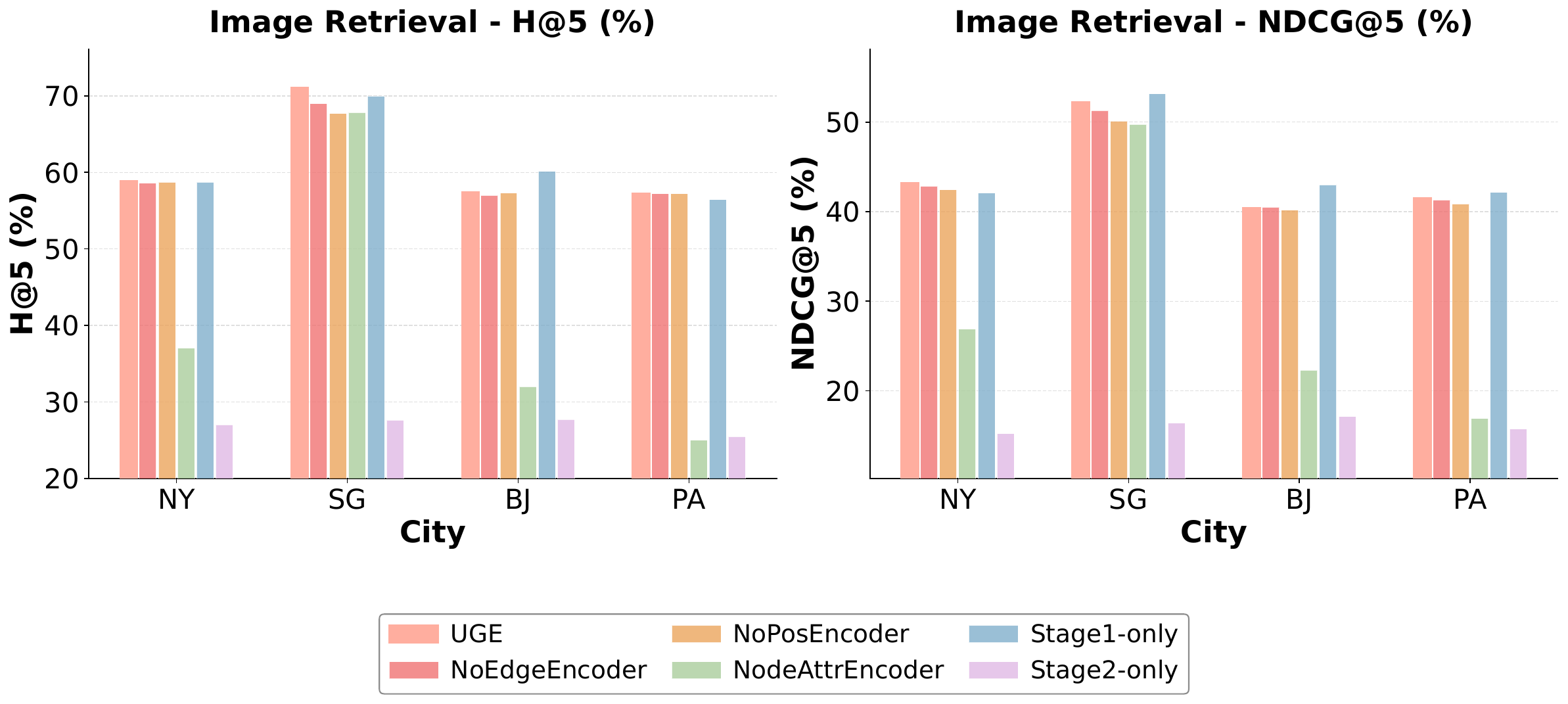}
\captionof{figure}{Results of ablation study on \emph{Image Retrieval}.}
\label{fig:ablation_image_retrieval}
\end{center}

\begin{center}
\captionsetup{type=figure}
\includegraphics[width=\columnwidth,height=0.28\textheight,keepaspectratio]{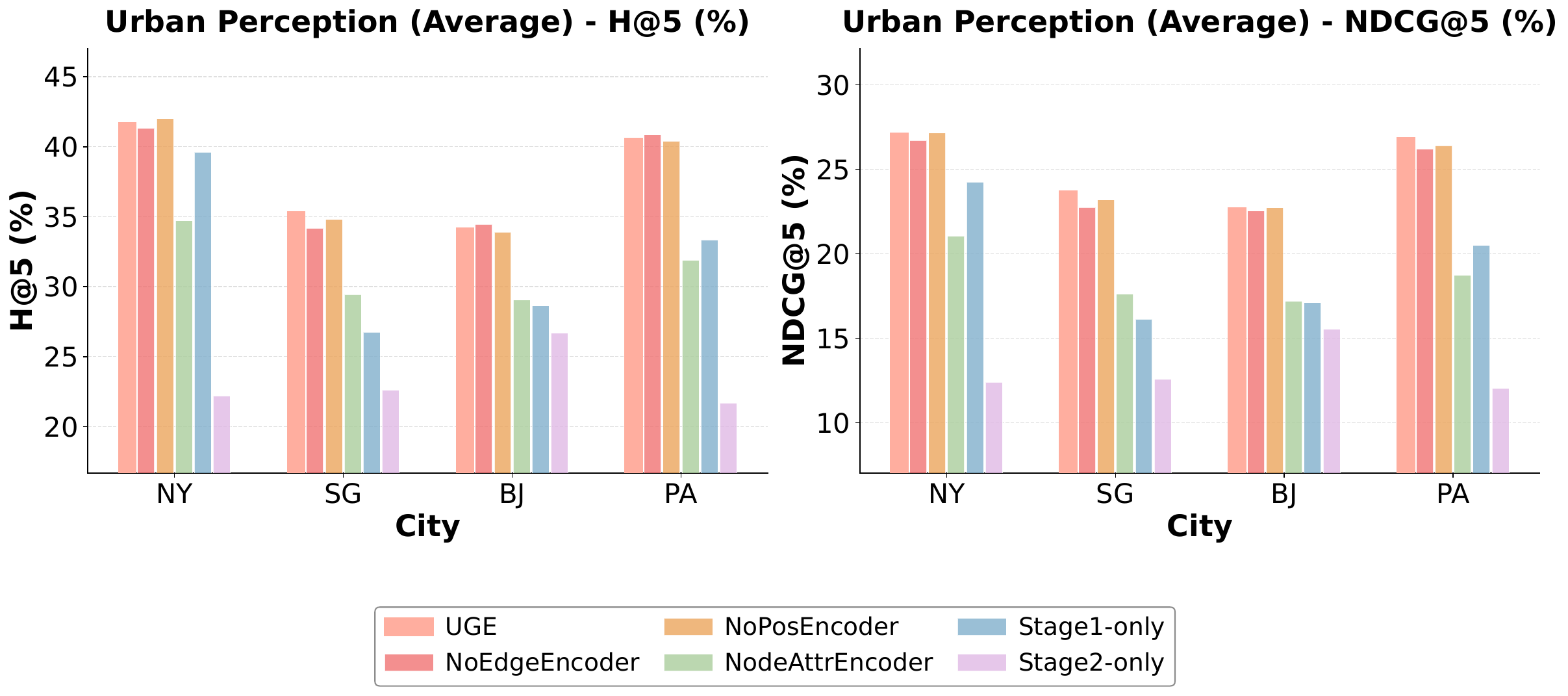}
\captionof{figure}{Average results of ablation study on \emph{Urban Perception}.}
\label{fig:ablation_urban_perception}
\end{center}

\begin{center}
\captionsetup{type=figure}
\includegraphics[width=\columnwidth,height=0.28\textheight,keepaspectratio]{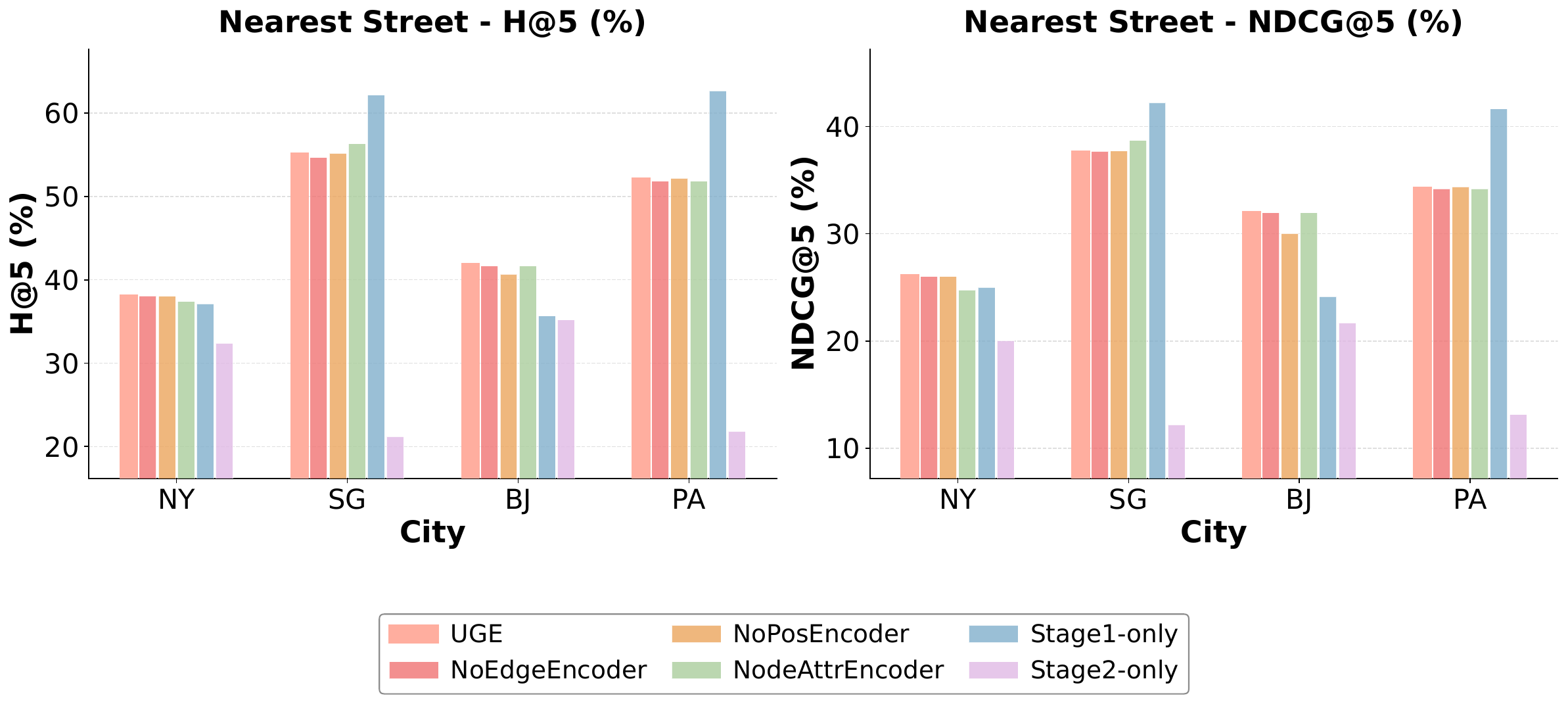}
\captionof{figure}{Results of ablation study on \emph{Nearest Street}.}
\label{fig:ablation_nearest_street}
\end{center}

\begin{center}
\captionsetup{type=figure}
\includegraphics[width=\columnwidth,height=0.28\textheight,keepaspectratio]{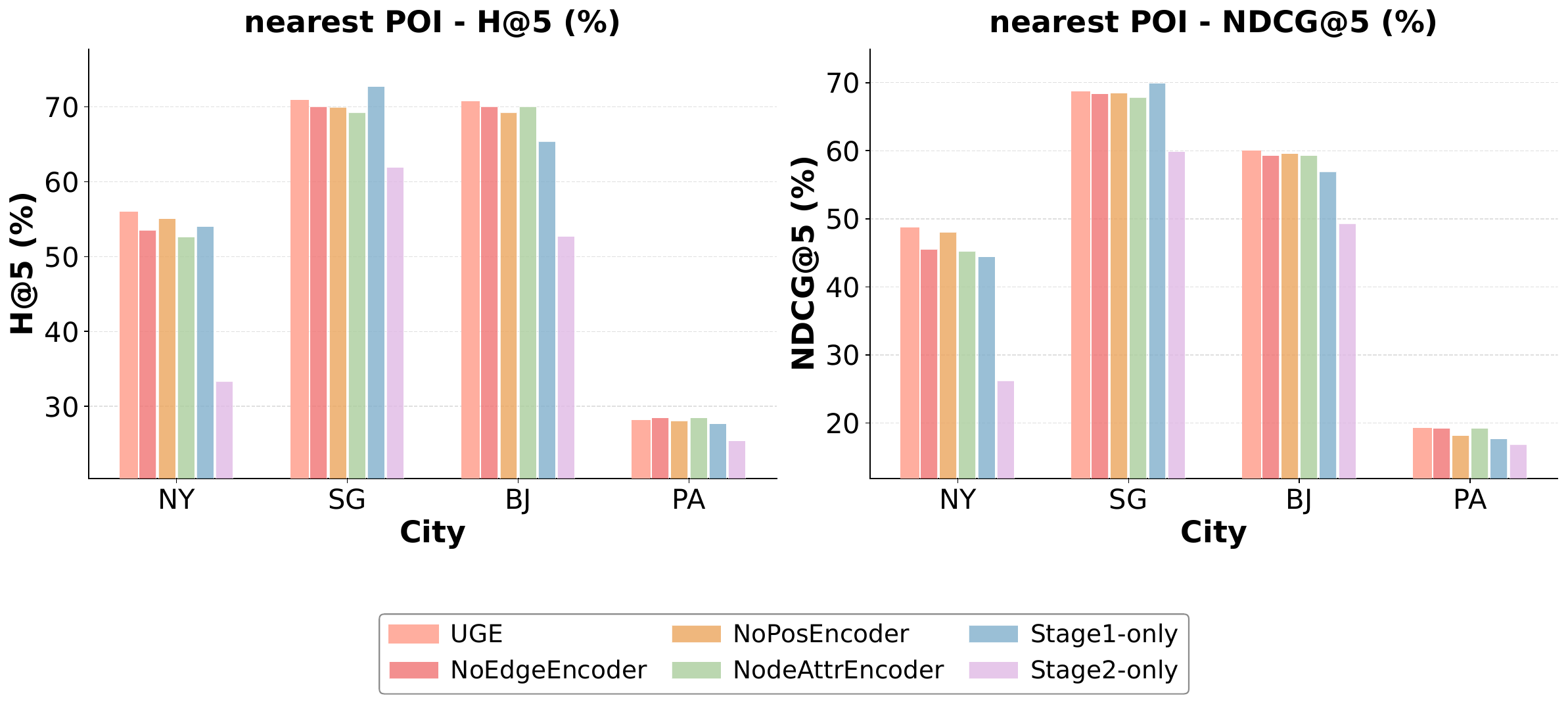}
\captionof{figure}{Results of ablation study on \emph{Nearest POI}.}
\label{fig:ablation_nearest_poi}
\end{center}

\begin{center}
\captionsetup{type=figure}
\includegraphics[width=\columnwidth,height=0.28\textheight,keepaspectratio]{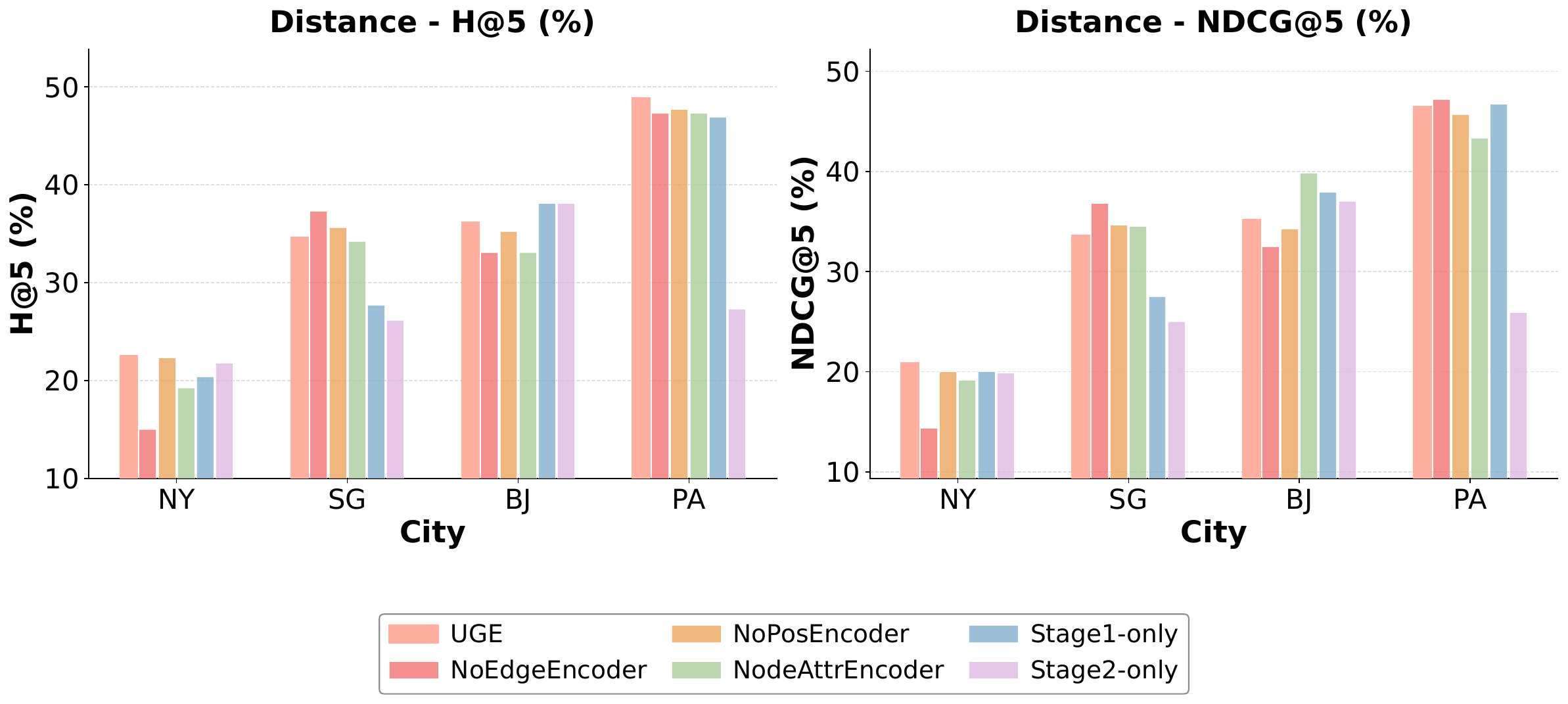}
\captionof{figure}{Results of ablation study on \emph{Distance}.}
\label{fig:ablation_distance}
\end{center}

\begin{center}
\captionsetup{type=figure}
\includegraphics[width=\columnwidth,height=0.28\textheight,keepaspectratio]{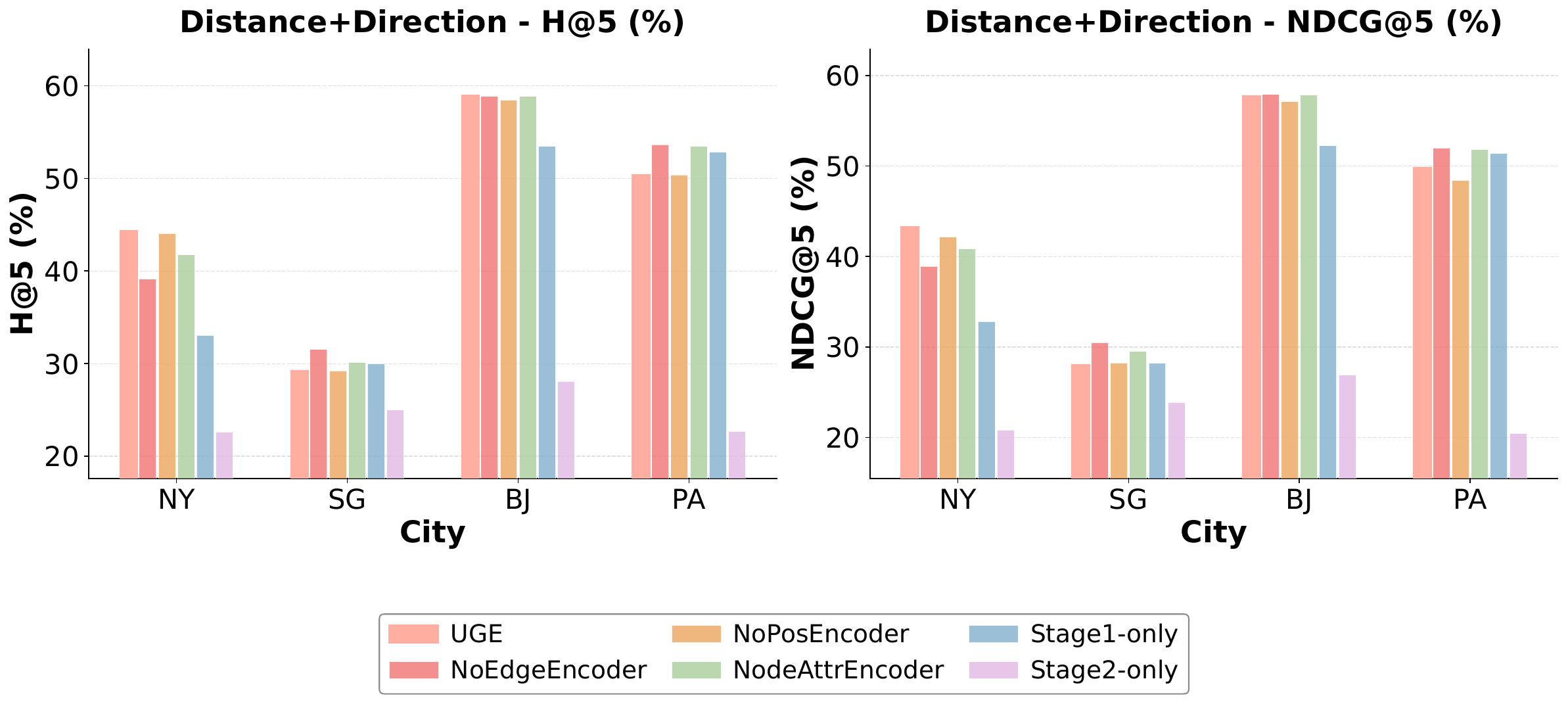}
\captionof{figure}{Results of ablation study on \emph{Distance-Direction}.}
\label{fig:ablation_distance_direction}
\end{center}

\subsection{Hyperparameter Study}
\label{app:hyper}

We study two key hyperparameters: the edge embedding dimension of the graph encoder and the learning-rate ratio between the Stage-2 graph encoder and LoRA adapters.
As shown in Figure~\ref{fig:edge_dim}, an edge embedding dimension of 64 consistently yields the best performance.
Figure~\ref{fig:lr_scale} further shows that controlling updates to the image--text alignment is critical for stable spatial knowledge injection, with smaller learning-rate ratios generally improving representation quality, while image retrieval benefits from stronger joint learning among graph, image, and text.

\begin{center}
\captionsetup{type=figure}
\includegraphics[width=\columnwidth,height=0.28\textheight,keepaspectratio]{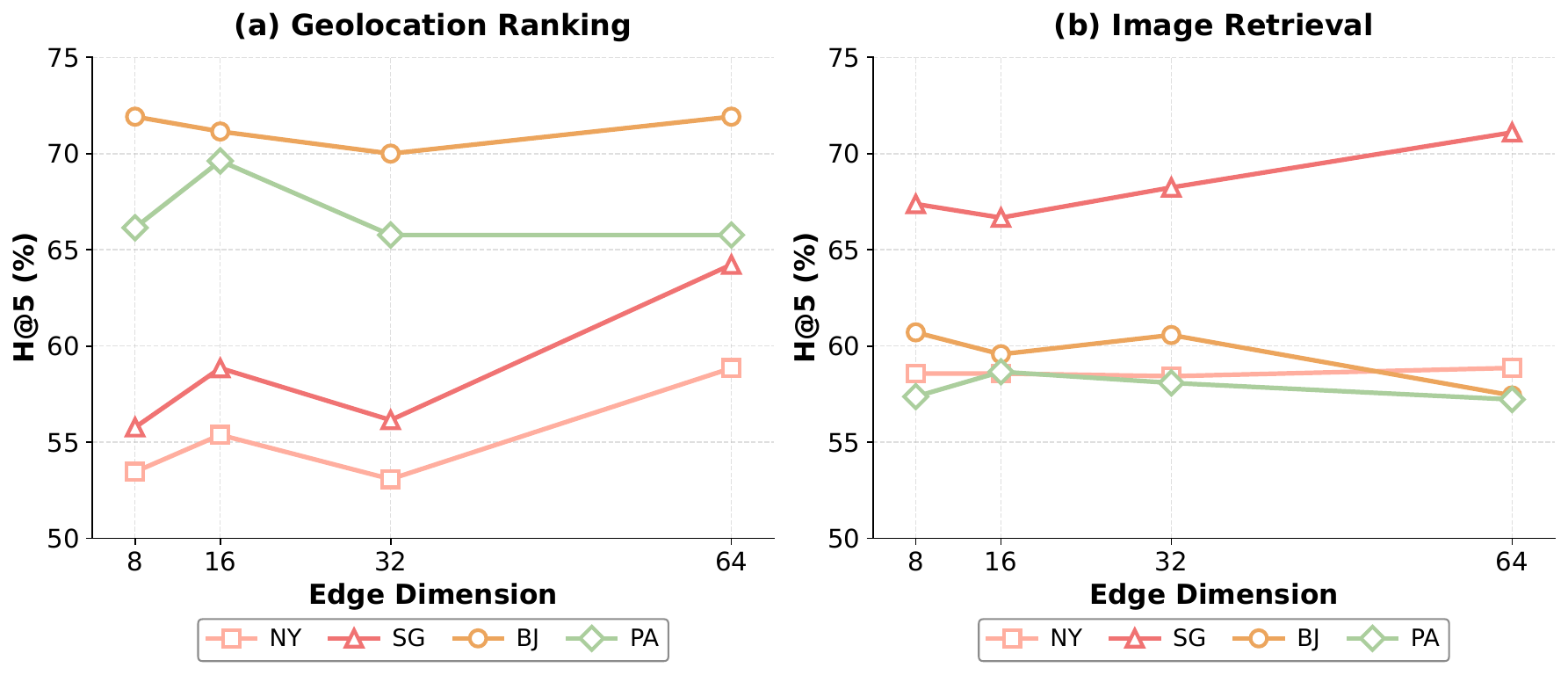}
\par\vspace{0.3em}
\includegraphics[width=\columnwidth,height=0.28\textheight,keepaspectratio]{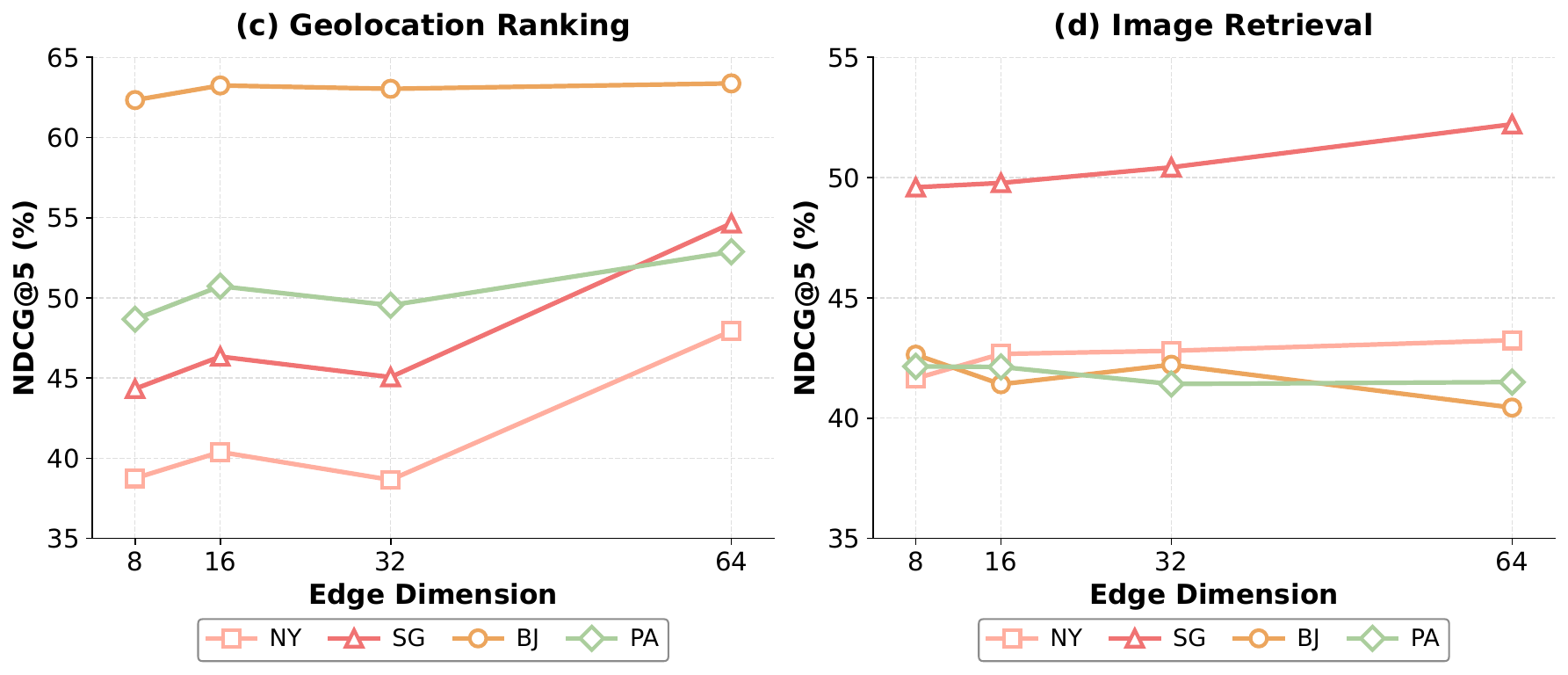}
\captionof{figure}{Effect of edge embedding dimension in the graph encoder.
(a,b) Hit@5 on geolocation ranking and image retrieval.
(c,d) NDCG@5 on geolocation ranking and image retrieval.}
\label{fig:edge_dim}
\end{center}

\begin{center}
\captionsetup{type=figure}
\includegraphics[width=\columnwidth,height=0.28\textheight,keepaspectratio]{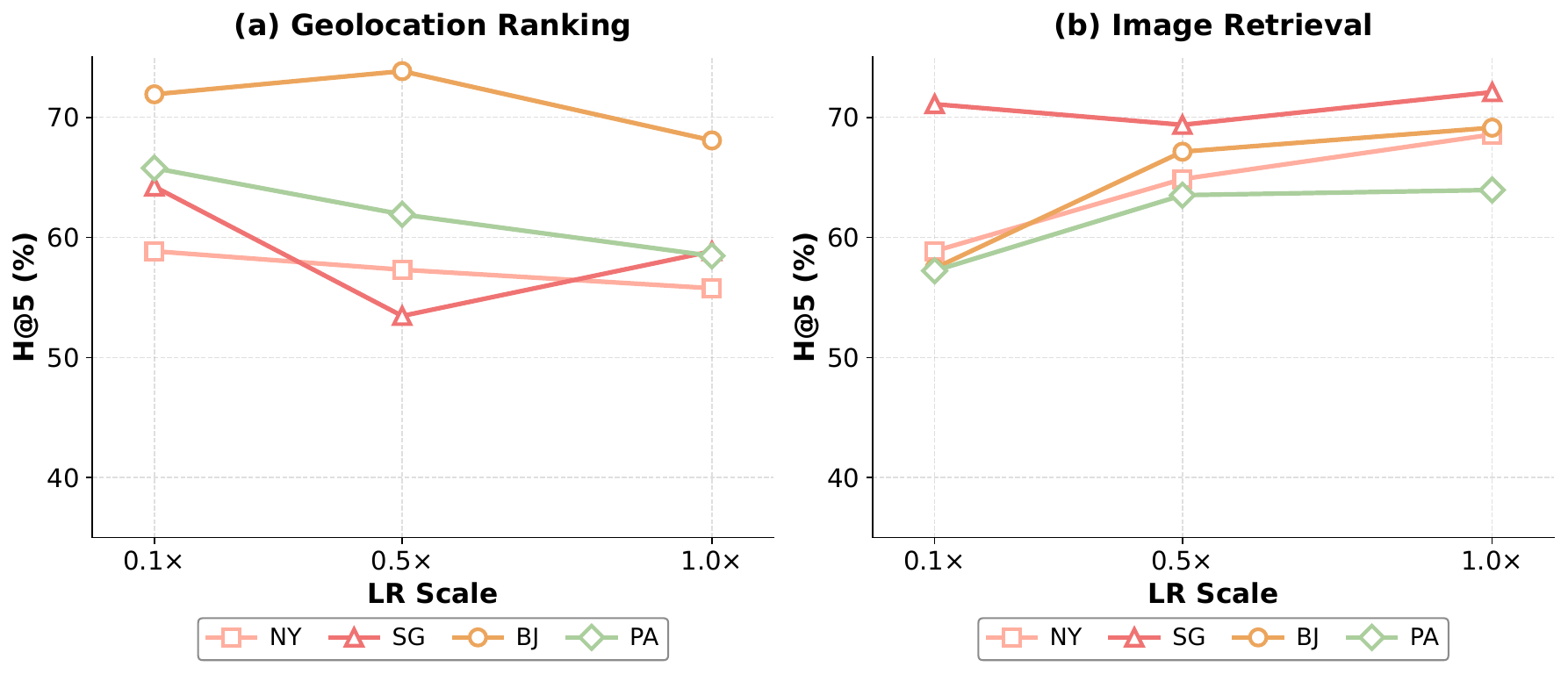}
\par\vspace{0.3em}
\includegraphics[width=\columnwidth,height=0.28\textheight,keepaspectratio]{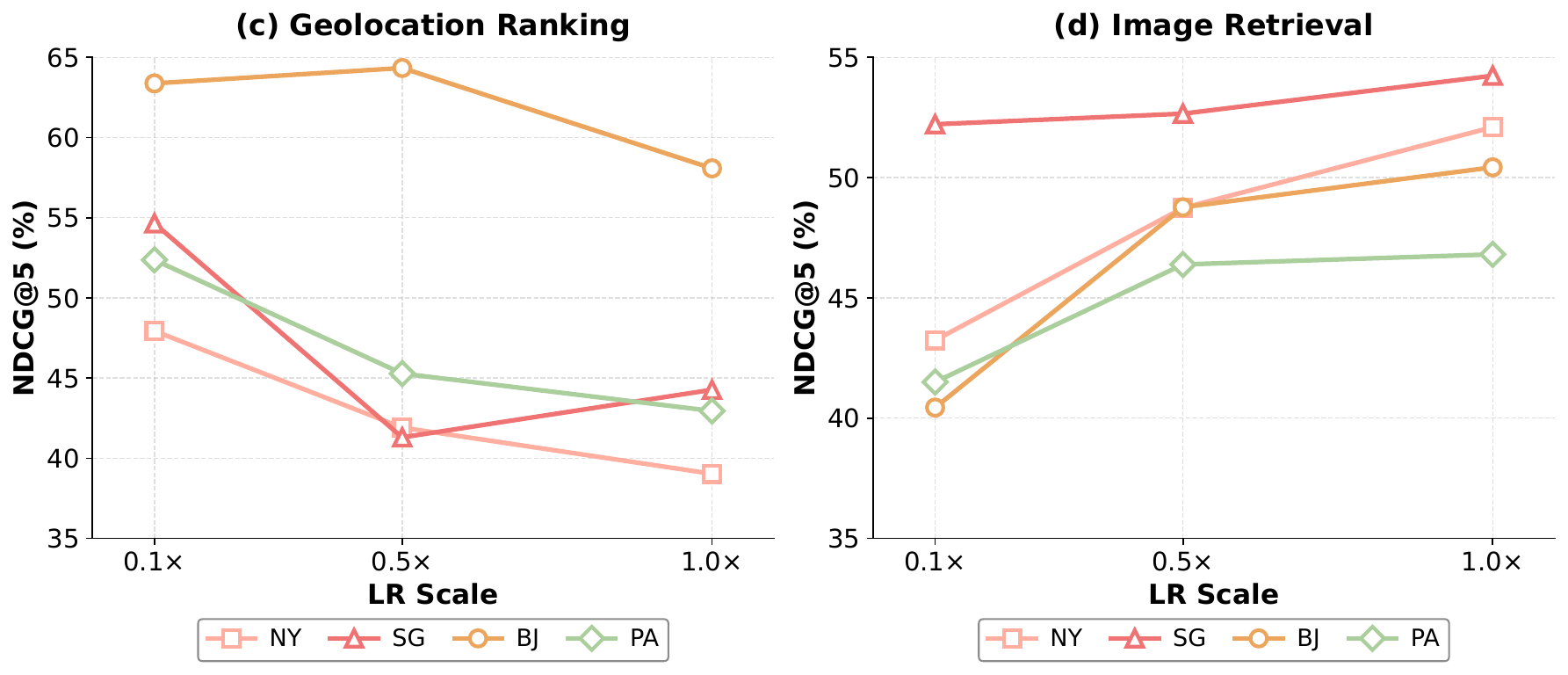}
\captionof{figure}{Effect of learning-rate scale between the Stage-2 graph encoder and LoRA adapters.
(a,b) Hit@5 on geolocation ranking and image retrieval.
(c,d) NDCG@5 on geolocation ranking and image retrieval.
Smaller LR scales generally yield more stable improvements, while larger scales favor stronger graph--text coupling for retrieval.}
\label{fig:lr_scale}
\end{center}

  %
\subsection{Inference efficiency} We report per-sample inference time on image retrieval. Graph conditioning increases latency from 2.9 s (image-only) to 11.0 s due to graph encoding and fusion.
\end{document}